# Time-Series Pattern Recognition in Smart Manufacturing Systems: A Literature Review and Ontology

Mojtaba A. Farahani[a,1], M. R. McCormick[a,1], Robert Gianinny[a], Frank Hudacheck[a], Ramy Harik[b], Zhichao Liu[a] and Thorsten Wuest[a]

[a]*West Virginia University, Morgantown, 26505 WV, U.S.A*
[b]*University of South Carolina, Columbia, 29208, SC, U.S.A*

**ABSTRACT**

Since the inception of Industry 4.0 in 2012, emerging technologies have enabled the acquisition of vast amounts of data from diverse sources such as machine tools, robust and affordable sensor systems with advanced information models, and other sources within Smart Manufacturing Systems (SMS). As a result, the amount of data that is available in manufacturing settings has exploded, allowing data-hungry tools such as Artificial Intelligence (AI) and Machine Learning (ML) to be leveraged. Time-series analytics has been successfully applied in a variety of industries, and that success is now being migrated to pattern recognition applications in manufacturing to support higher quality products, zero defect manufacturing, and improved customer satisfaction. However, the diverse landscape of manufacturing presents a challenge for successfully solving problems in industry using time-series pattern recognition. The resulting research gap of understanding and applying the subject matter of time-series pattern recognition in manufacturing is a major limiting factor for adoption in industry. The purpose of this paper is to provide a structured perspective of the current state of time-series pattern recognition in manufacturing with a problem-solving focus. By using an ontology to classify and define concepts, how they are structured, their properties, the relationships between them, and considerations when applying them, this paper aims to provide practical and actionable guidelines for application and recommendations for advancing time-series analytics.

## 1. Introduction

We are in the midst of a new era in manufacturing driven by the implementation of Industry 4.0 infrastructure, smart manufacturing technologies, and the mass digitalization of assets and processes across industries. This digital transformation toward Smart Manufacturing Systems is built upon the three key principles of connectivity, virtualization, and data utilization [131]. It is enabled by data-driven analytics tools like Artificial Intelligence (AI), Machine Learning (ML), digital twins, cloud computing, and big data analytics [151][146][66]. These rapidly developing technologies, coupled with the exponential increase in available manufacturing data, have enabled the execution of data-hungry AI and ML problem-solving methods at an unprecedented scale [53].

Time-series analytics is a tool enabled by these technologies, and related data is used across a wide variety of industries such as meteorology, financial markets, healthcare, and manufacturing. In manufacturing, time-series data is utilized across both Operational Technology (OT) and Information Technology (IT) to generate outcomes such as improved efficiency, quality, and performance, as well as reduced waste. By utilizing data-driven analytics tools in smart manufacturing systems, hidden insights can be extracted from time-series data in ways not achievable through traditional analysis methods [77].

There are two classes of input dimensionality in time-series analytics: univariate and multivariate. A univariate time-series $x = [x_1, x_2, ..., x_T]$ is an ordered vector of real values with length $T$ [31]. A M-dimensional multivariate time-series, $X = [x^1, x^2, ..., x^M]$ is a matrix consists of $M$ different univariate time-series with $x^i \in \mathbb{R}^i$. A dataset $D = \{(X_1, Y_1), (X_2, Y_2), ..., (X_N, Y_N)\}$ is a collection of $(X_i, Y_i)$ pairs where $X_i$ could either be a univariate or multivariate time-series and $Y_i$ is the corresponding supervisory label. The pattern recognition approach is either supervised or semi-supervised if label value $Y_i$ are available and used, and unsupervised if label values $Y_i$ are not used or are not available. A visual representation of time-series dimensions are shown in Figure 1.

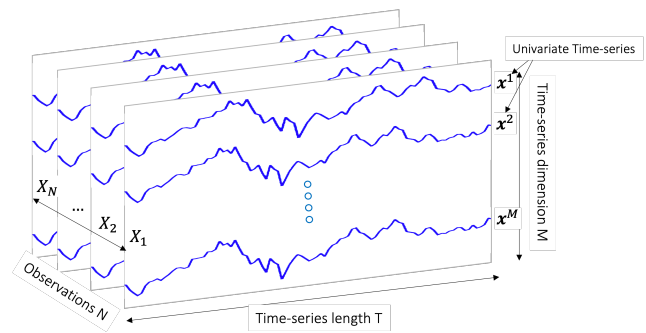

**Figure 1:** Time-Series Definition

✉ ma00048@mix.wvu.edu (M.A. Farahani); mtmccormick@mix.wvu.edu (M.R. McCormick); harik@cec.sc.edu (R. Harik);
zhichao.liu@mail.wvu.edu (Z. Liu); thwuest@mail.wvu.edu (T. Wuest)
ORCID(s):
[1]Equal Contribution, Co-First Author

ML has been widely used across many domains to provide advanced analysis capabilities. This includes applications such as computer-vision systems assisting human operators in monitoring the quality of products, automated real-time fault and event detection systems, social media pattern recognition [42], autonomous systems [138], and Remaining Useful Life (RUL) prediction of manufacturing tools [134]. These capabilities and outcomes provide tangible value to manufacturers who require a justifiable return on investment to consider investigating and pursuing these smart manufacturing technologies. However, even with a justification for investment, practical guidance on their implementation is not readily available or accessible. In addition, academic literature spans a wide range of disciplines and applications without a cohesive roadmap. To address this manufacturing-specific research gap, and while detailed literature reviews on the topic of time-series *forecasting* have already been conducted, we conducted a systematic literature review on the underrepresented topic of time-series *pattern recognition*. In doing so, we distilled the various concepts to a practical level, organized them using an ontology, and provided the tools necessary to navigate the topic. Given the rapidly increasing popularity of time-series pattern recognition illustrated in Figure 2 which is derived from the 1254 papers in the Database Search section, this review is needed now more than ever.

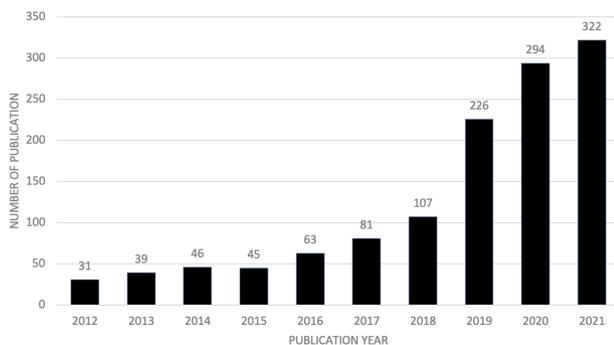

Figure 2: Database Search Results

### 1.1. Related Reviews

The topic of ML in manufacturing has been discussed extensively over the last 30 years. Multiple literature reviews provided high-level overviews of ML applications in manufacturing. Wuest et al. discussed the suitability, advantages, and challenges of using ML techniques in manufacturing [151]. Bertolini et al. conducted a review on ML algorithms for industrial applications in the operations management field, focusing on multiple application domains including Maintenance Management, Quality Management, Production Planning and Control, and Supply Chain Management [9]. Hegde et al. reviewed ML methods for engineering risk assessment encompassing identification, analysis, and evaluation across industries [45]. Nti et al. presented a comprehensive review of studies on AI and its applications in engineering and manufacturing, capturing the progress of ML methods in various engineering and manufacturing environments. While their meta-literature review summarized 18 previous literature review papers which focused on a wide variety of application areas, they concluded that they had not adequately addressed the breadth of the progress of ML in engineering and manufacturing, and suggested that there is a need for an organized, comprehensive, systematic literature review presenting that progress [108].

In tandem with high-level overviews, other review papers provided a focused analysis of specific applications and industries. While there are too many to highlight individually, the following examples showcase the application-specific nature of these available reviews. Mata et al. focused on optical networks including performance monitoring and characterization of network components, and transmission quality estimation [92]. Antonopoulos et al. illuminated energy demand-side response including consumer characterization and selection optimization with schedule forecasting with dynamic pricing and control [3]. Babic et al profiled image-based quality inspection techniques, providing a cross-section of the benefits, challenges, and application areas for various approaches [6]. These papers provided in-depth analysis and insights tailored to addressing application-specific challenges. In contrast, this paper focuses specifically on the topic of time-series pattern recognition in manufacturing.

### 1.2. Motivation

While prior works provided both high-level overviews and focused analysis for specific applications and industries, none focused specifically on the topic of time-series pattern recognition in manufacturing. This paper is intended to close this research gap and enable readers to better understand the landscape of problems regularly addressed using time-series analytics, as well as identify ML methods that can be used to successfully solve them in manufacturing settings. In addition, this paper is intended to illuminate and demystify distinctions regarding ML which are not readily perceivable to readers with limited backgrounds in this growing field, such as manufacturing engineers in industry. We strive to provide a resource to the reader that enables them to identify high-level methods for solving problems, identify challenges in implementing solutions, understand what other practitioners are using to solve similar problems, and identify publications where more detailed information can be found on specific applications for their implementation journey.

### 1.3. Scope

The scope of this systematic literature review encompasses ML applications of time-series pattern recognition in manufacturing and engineering settings since the inception of industry 4.0 in 2012 through the lens of and with a focus on solving manufacturing problems. Using an "include unless not applicable" philosophy, literature was limited to established, high-quality publications relating to manufacturing where ML was applied in a time-series pattern recognition context. In addition, purely theoretical or conceptual papers without practical application in manufacturing were excluded to ensure actionable guidance toward application for the target audience.

During the review process and especially during the data extraction step, challenges were encountered with respect to the consistency and depth of information provided by the literature. Rather than attempting to analyze attributes based on partial data derived from fractions of papers, we decided to limit the scope of analysis to attributes which were consistently found in the literature and provide recommendations on what information would be useful during future investigations. This was driven by the perception of a limited return on investment in comparison to what could be achieved through greater efficiency in the future through adherence to recommendations developed by this literature review.

### 1.4. Organization

The remainder of this paper is organized as follows: The Literature Review Methodology section describes the research methodology of this literature review that led to the selection of and data extraction from reviewed papers. Based on the insights gained from the analysis of selected literature, we share and discuss the results in the Results & Discussion section. To extend the Results and Discussion section and to facilitate application, actionable guidance is provided in the Guidelines section. Finally, conclusions drawn from the previous sections and recommendations for future research are presented in the Conclusions section.

## 2. Literature Review Methodology

The objective of the literature review was to i) identify relevant peer-reviewed journal articles and conference papers that cover the target subject matter, and ii) extract data and information that contributes to achieving our paper's objectives. We followed a rigorous and transparent five-step process of 1) database search, 2) title evaluation, 3) abstract evaluation, 4) detailed evaluation, and 5) data extraction. Each step is described in more detail in their respective subsections. The five-step process and respective quantity of papers remaining after each step are illustrated in Figure 3. In each evaluation step, we defined three criteria and all evaluations were performed based on exclusion rather than inclusion. In this way, the papers outside of the defined scope were discarded and the remaining papers were kept for reevaluation in succeeding steps.

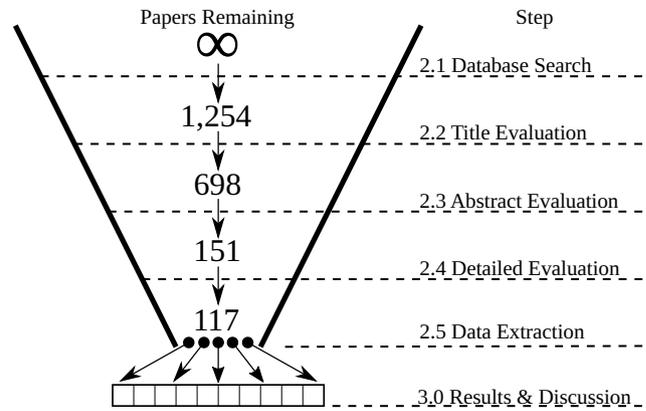

**Figure 3:** Literature Review Funnel

### 2.1. Database Search

A keyword-based search was performed using the SCOPUS database to generate a pool of articles. SCOPUS is a well-known, comprehensive database in the field of engineering, and is considered one of the best options for high-quality academic literature. Since Industry 4.0 and related publications originated in 2012 (one year after the term was coined), we only considered publications starting from 2012. The database search (step 1) using the search string shown in Figure 4 yielded 1,254 papers. The main query criteria are illustrated in Figure 5. Further, Figure 2 showcases the distribution over time derived from this section.

TITLE-ABS-KEY ( ( ai OR ml OR "Artificial intelligence" OR "Machine Learning" OR "Deep Learning" OR "ANN" OR "Artificial Neural Networks" ) AND ( "Time series" OR "Time series analysis" OR "Pattern" OR "Pattern Recognition" OR "event detection" ) AND ( manufacturing OR production ) ) AND ( LIMIT-TO ( DOCTYPE , "ar" ) OR LIMIT-TO ( DOCTYPE , "cp" ) ) AND ( LIMIT-TO ( SUBJAREA, "ENGI" ) ) AND ( LIMIT-TO ( PUBYEAR , 2021 ) OR LIMIT-TO ( PUBYEAR , 2020 ) OR LIMIT-TO ( PUBYEAR , 2019 ) OR LIMIT-TO ( PUBYEAR , 2018 ) OR LIMIT-TO ( PUBYEAR , 2017 ) OR LIMIT-TO ( PUBYEAR , 2016 ) OR LIMIT-TO ( PUBYEAR , 2015 ) OR LIMIT-TO ( PUBYEAR , 2014 ) OR LIMIT-TO ( PUBYEAR , 2013 ) OR LIMIT-TO ( PUBYEAR , 2012 ) ) AND ( LIMIT-TO ( LANGUAGE , "English" ) )

**Figure 4:** Database Search String

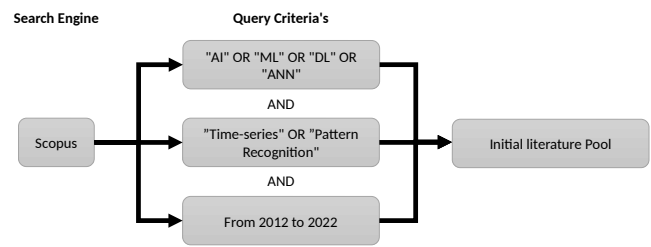

**Figure 5:** Database Search Flowchart

### 2.2. Title Evaluation

The titles (step 2) of all 1,254 papers returned by the database search were evaluated based on the following exclusion criteria with the intent of excluding content that is outside of the scope defined in the Scope section.

1. Titles not related to engineering or manufacturing or production were excluded.
2. Titles not related to time-series analytics or pattern recognition were excluded.
3. Titles that did not use ML techniques were excluded.

After the completion of the title evaluation (step 2), 556 papers were evaluated as being outside the scope and were discarded. For instance, the title of Mia et al. in [99] implied the use of time-series data but lacked a focus within the manufacturing setting. Likewise, the title of Xie et al. in [153] implied the use of ML pattern recognition but also lacked a focus within the manufacturing setting. In addition, the title of Suga et al. in [135] did not imply the use of ML concepts as well as no focus on a manufacturing or industrial setting.

### 2.3. Abstract Evaluation

The abstracts (step 3) of the 698 remaining papers after the title evaluation were evaluated based upon the following criteria:

1. Abstracts that did not indicate the use of pattern recognition were excluded.
2. Abstracts that did not indicate the use of time-series data were excluded.
3. Abstracts that did not indicate practical industrial applications, with only a theoretical or conceptual basis, were excluded.

Once the abstract evaluation was completed, 547 papers were found to not align with the scope of this literature review and were discarded. For instance, the abstract of Fukuda et al. in [35] contained information regarding pattern recognition that was used for image recognition and did not utilize any type of time-series data. Zheng et al. in [164] discussed ML concepts in a manufacturing setting but lacked the use of time-series data. Ren et al. in [117] covered topics in a manufacturing setting but lacked the use of both time-series data and pattern recognition.

During the abstract evaluation step, two insights emerged which were noteworthy. First, many pattern recognition papers which did not use time-series data were focused on image pattern recognition (Image PR), which was summarized in a literature review by Babic et al. [6]. Second, many papers that did not focus on pattern recognition but included time-series data were focused on time-series forecasting. No similar insights emerged from non-pattern recognition, non-time-series focused papers. The results of the abstract evaluation are shown in Table 1.

### 2.4. Detailed Evaluation

The remaining 151 papers after the abstract evaluation were evaluated (step 4), reviewed in detail, and evaluated for exclusion based on the following criteria:

1. Papers that did not indicate the use of pattern recognition were excluded.
2. Papers that did not include the use of time-series data were excluded.
3. Papers that did not indicate practical industrial applications, with only a theoretical or conceptual basis, were excluded.

After the detailed review, 34 papers were evaluated as being outside of scope and were discarded. For example, Liu et al. in [84] discussed the use of time-series forecasting as well as ML, but did not apply these concepts in a manufacturing setting. Butt et al. in [12] focused on ML topics but did not include time-series data nor was it set in the manufacturing domain. Khanduja in [57] focused on time-series data yet did not connect it to ML applications. Since feature extraction by itself is not considered a type of pattern recognition, papers that solely focused on algorithms used for feature extraction were also excluded. For example, Susto et al. in [136] proposed Supervised Aggregative Feature Extraction for Big Data Time-Series Regression.

### 2.5. Data Extraction

After the title, abstract, and detailed evaluation of the papers, data was extracted from the remaining 117 papers for quantitative analysis. Some attribute values were extracted on a per paper basis, such as Input Dimensionality, Industry, and Problem. For other attributes, multiple attribute values were extracted from a single paper, such as Data Sources and Algorithms. Attributes such as Approach and Technique were classified on a per Algorithm basis, as detailed in the Algorithm Classification section.

In order to contextualize extracted data, the method by which each attribute was extracted and the chain of custody of attribute values is outlined in Table 2. For papers which examined multiple algorithms, all algorithms were included, including cases where established algorithms were compared to new theoretical algorithms.

|  |  | Pattern Recognition | |
|---|---|---|---|
|  |  | Yes | No |
| Time Series | Yes | 151 Included | 202 Excluded (Forecasting) |
|  | No | 246 Excluded (Image PR) | 99 Excluded |

**Table 1:** Abstract Evaluation Matrix

| Extraction Method | Attributes |
|---|---|
| 1. Attributes extracted directly from database search results. | Author Keywords, Index Keywords |
| 2. Attributes extracted directly from the paper content during detailed evaluation and left in the author's original terminology. | Algorithm |
| 3. Attributes extracted either directly from or inferred from the paper content during detailed evaluation. | Data Source, Input Dimensionality, Industry |
| 4. Attributes determined through analysis and classification of Author Keywords, Index Keywords, Title, and Abstract. If the attribute value was not clear in any of these sources, the attribute value was derived from paper content. | Problem |
| 5. Attributes determined through a detailed analysis of paper content and left in the author's original terminology. | Jargon |
| 6. Attributes determined through a detailed analysis of paper content and mapped to our ontology. | Solution, Approach, Technique |

**Table 2:** Data Extraction Methods

## 3. Results & Discussion

In this section, the results formulated from the data extraction process are presented. While traditional literature review methods emphasize separating results and discussion into distinct sections to avoid biasing the results with commentary, we found that combining results and discussion provides a better experience for the reader due to dependency on common figures and consolidation of complex ideas rather than distributing them over multiple sections and requiring extraneous page flipping.

This section begins with presenting an ontology consisting of a problem domain and a solution domain which provide a structure for understanding the results. Then, results pertaining to the problem domain are presented, followed by results pertaining to the solution domain. Finally, the relationship between the problem and solution domains are examined.

### 3.1. Ontology

While performing the detailed evaluation, data extraction, and subsequent analysis of the extracted data, it became evident that a cohesive and consistent organizational structure for presenting and correlating concepts was not present in the examined literature. In addition, the terminology was inconsistently used across the reviewed papers, which consistently appeared in four different forms:

- Different authors used the same terminology to describe different concepts.
- Different authors used different terminology to describe the same concept.
- Different authors classified or structured concepts differently based on the same definitions.
- Different authors classified or structured concepts differently based on differing definitions.

To bridge this gap, a system that provides a consistent structure using consistent terminology was necessary in order to compare, contrast, and clearly convey concepts. In addition, an easy to understand system for providing practical and actionable guidance with a contextual map of the landscape was needed. By envisioning the perspectives and needs of both the researchers producing these works and the academic and industry professionals who will consume and apply the content of this paper, we formulated an ontology to provide a concise visual roadmap to facilitate solving manufacturing problems using time-series analytics. The ontology was not formulated from the content of the reviewed papers, but rather as an attempt to make sense of and apply their content. In essence, the ontology is both a result and extension of the analysis of the extracted data.

Throughout the remainder of this paper, capitalization is used to signify terminology specific to the proposed ontology, whereas lower case is used to signify the colloquial use of the term. Alternative terms are used to avoid confusion whenever possible, but this syntax is used to distinguish between cases wherever possible.

#### 3.1.1. Problem and Solution Domains

At its core, each of the reviewed papers attempted to resolve a Problem by applying a single Solution or a combination of Solutions. When limiting the domain scope to technical applications of time-series analytics rather than high level business objectives (e.g., Return on Investment Analysis), the *problem statement* describes the transformation of data (inputs) into desired insights (outputs), as well as the relationship between the data and insights, without defining how that transformation takes place. In tandem, a *Solution* defines the transformation of data into insights without the context of what that transformation or the resulting insights signify, and could theoretically be applied to multiple unique Problems. The domains which comprise the ontology are illustrated in Figure 6. The process of acquiring the data and the decisions made based on the insights obtained are outside the scope of and define the boundary of the ontology.

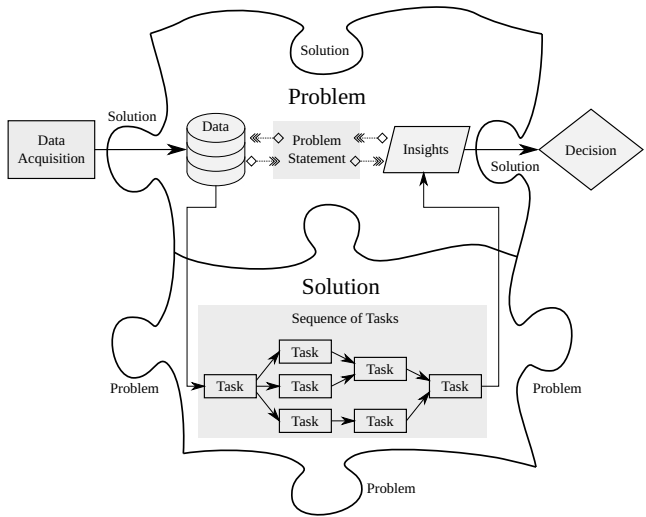

**Figure 6:** Domains

Since the data and insights are the only entities crossing the domain boundary, they in effect define the compatibility between a Problem and Solution, as caricatured by the compatibility of puzzle pieces. There are few elements defining compatibility, which enables flexible substitution. As a result, a single Problem can be resolved by multiple unique Solutions given that each Solution is compatible with the Problem. The identification and illustration of compatibility and interchangeability were objectives when formulating the ontology.

While a single Solution could theoretically be applied to multiple unique Problems, we did not see any examples of this in the reviewed papers and found the reviewed papers to be problem-driven rather than solution-driven. However, the applicability of similar Solutions to multiple Problems may become a factor in solution-driven niche areas such as resource-constrained embedded devices where a narrow range of Solutions are feasible. In short, the interchangeability and compatibility of Problems and Solutions are solely defined by the data and insights regardless of whether a use case is Problem or Solution driven.

The definitions employed throughout the ontology are derived from a combination of the Oxford English Dictionary, the Merriam-Webster dictionary, common use in ML literature, and strategic embellishment by the authors of this paper, as illustrated in Table 3.

| Term | Definition |
|---|---|
| Problem | A matter or situation that needs to be dealt with or overcome through transforming data into insights which enable decision making. |
| Solution | A means or method of resolving a problem. |
| Task | A unit of work to be done or undertaken which produces a specific result. |
| Data | Facts and information which do not provide a deep intuitive understanding of their meaning and significance. |
| Insights | Facts and information which provide a deep intuitive understanding of their meaning and significance. |

**Table 3:** Ontology Definitions

The structure of a Solution consists of a sequence of *tasks*, each of which is a unit of work to be undertaken. As shown in Figure 7, a sequence of tasks can be constructed such that the output data or insights of one task can be used as the input of another task, and the insights of multiple tasks can be combined to determine global insights.

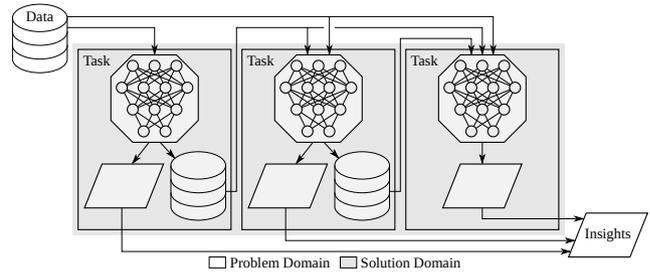

**Figure 7:** Sequence of Tasks in a Solution

The structure of a task is distinctly different from a sequence of tasks, as illustrated in Figure 8. A task can be composed of a single or multiple Algorithms which function as a single Algorithm, and cannot be decomposed like a sequence of tasks. This is the distinguishing feature of a task that drives functional differences between a task and a sequence of tasks. While tasks can be executed independently, the Algorithms within a task must be executed (and likely trained) as a unit. Likewise, a downstream task cannot be executed until an upstream task has been completed, while a task is self-contained and algorithms within it can be executed in a combination of parallel or series. In addition, tasks are not inherently constructed from ML algorithms and can perform rudimentary computational operations such as pre-processing or transformations during which traditional algorithms are used. Since the focus of this paper is ML methods, figures and commentary primarily focus on ML.

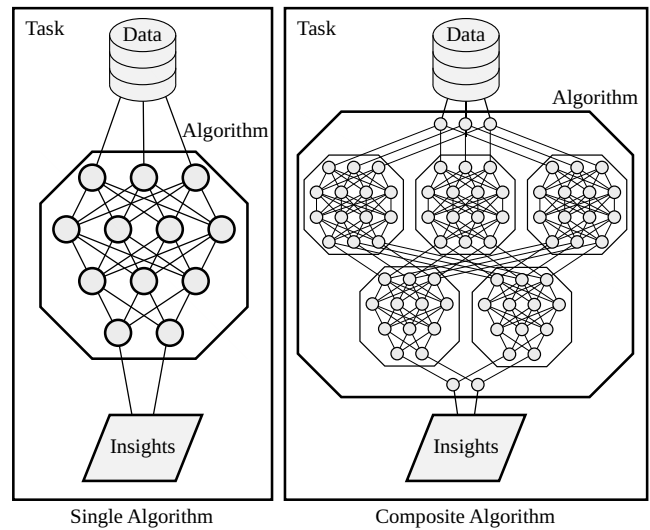

**Figure 8:** Structure of a Task

### 3.1.2. Algorithm Classification

To guide decision-making and Algorithm selection, it is useful to compare and contrast Algorithms through classification. Due to the diversity of research and publications regarding ML across a multitude of sub-domains, industries, and applications, a variety of methods for classifying ML algorithms have emerged and resulted in a lack of consensus [151]. This has presented a challenge for effectively analyzing the relationships between and structure of ML algorithms and drawing conclusions from related literature, as well as efficiently selecting algorithms based on that understanding. To overcome these challenges, we formulated a method of classification specific to time-series analytics in manufacturing to enable simplified analysis, highlight current challenges, and provide actionable guidance without the baggage and biases incurred from other sub-domains and applications. The structure of classification is illustrated in Figure 9 and the class definitions are presented in Table 4.

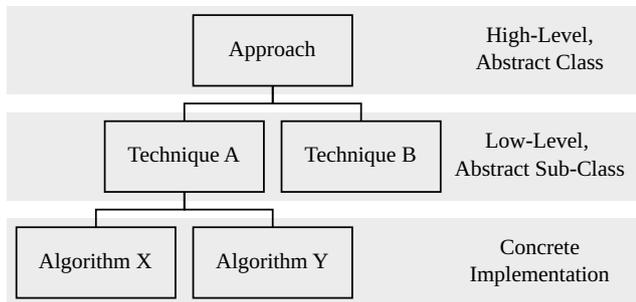

**Figure 9:** Algorithm Classification Tree

| Class | Definition |
| --- | --- |
| Approach | An abstract class of algorithms representing a method of executing a task which is defined by its output dimensionality, label requirements, and a correlation to task structure. |
| Technique | An abstract subclass of algorithms representing a method of executing a task that groups algorithms based on similarity. A technique may be used across multiple approaches or be unique to a specific approach. |
| Algorithm | A concrete implementation of a machine learning algorithm, which is a method of completing a task, and which can only belong to a single Technique. |

**Table 4:** Algorithm Classification Definitions

Two levels of classification were constructed to highlight the different characteristics of each Algorithm, as defined in Table 4. The "Approach" classes differentiate Algorithms based on objective differences which are easily identifiable and discernible, such as learning type, output dimensionality, and label requirements, as illustrated in Figure 9. The "Technique" subclasses group algorithms based on subjective similarities which are not as easily identifiable or discernible but directly impact Algorithm performance. By using both Approaches and Techniques, a practitioner should be able to efficiently identify a limited pool of applicable Algorithms by using the method proposed in the Algorithm Selection section.

Supervised Learning (SL), Semi-Supervised Learning (SSL), Unsupervised Learning (UL), and Reinforcement Learning (RL) are generally considered the primary approaches to ML. However, during the literature review process, no papers were identified which use a RL approach for time-series pattern recognition in manufacturing. This is supported by Wuest et al. who noted that RL is not yet widely used in manufacturing settings and there are few examples of use in manufacturing literature [151]. For this reason, RL is not included in Figure 10 or in the proposed ontology. SSL is a learning method distinct from and is a combination of SL and UL which warrants special consideration. This is due to functional differences and its ability to handle partially labeled data which is covered in more detail in the Guidelines section. Figure 10 provides a visual representation of Approaches based on learning type, output dimensionality, and label requirements.

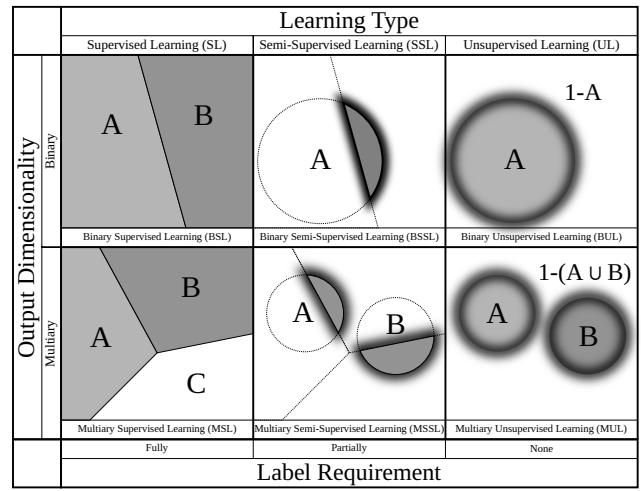

**Figure 10:** Approach Classification Matrix

Each Approach can be subdivided into a number of Techniques, and a given Technique can be employed under multiple Approaches as illustrated in Figure 11. However, each Algorithm can only belong to a single Technique, which is driven by the definition of a Technique being based on similarity and downstream intended function during Algorithm selection. Artificial Neural Networks (ANN) and Deep Learning (DL) are special interest Techniques due to their popularity and are distinct from Conventional ML Techniques with regard to structure. The Techniques listed in Figure 11 are for illustrative purposes only, are not a component of the ontology, and would be expected to change with the evolution of results in the future. However, the method of classification and selection of Techniques detailed in the Algorithm Selection section would likely remain the same, even with a shift toward objective rather than subjective Technique definitions.

|  | | Learning Type | | |
|---|---|---|---|---|
|  | | Supervised Learning (SL) | Semi-Supervised Learning (SSL) | Unsupervised Learning (UL) |
| Output Dimensionality | Binary | Instance-Based<br>Decision Tree-Based<br>Statistical<br>Regression | Semi-Supervised Detection of Outliers (SSDO) | Constraint-Based<br>Density-Based<br>Center-Based<br>Distribution-Based<br>Dimension Reduction<br>Connectivity-Based |
|  |  | Recurrent Neural Networks (RNN)<br>Convolutional Neural Networks (CNN)<br>Feed Forward Neural Networks (FFN)<br>Autoencoders (AE) |  | Autoencoders (AE)<br>Convolutional Neural Networks (CNN)<br>Recurrent Neural Networks (RNN)<br>Generative Adversarial Models<br>Feed Forward Neural Networks (FFN) |
|  | Multiary | Decision Tree-Based<br>Instance-Based<br>Statistical |  | Center-Based<br>Connectivity-Based<br>Density-Based<br>Dimension Reduction<br>Distribution-Based |
|  |  | Feed Forward Neural Networks (FFN)<br>Convolutional Neural Networks (CNN)<br>Recurrent Neural Networks (RNN)<br>Autoencoders (AE) |  | Generative Adversarial Models |
|  | | Fully | Partially | None |
|  | | Label Requirement | | |

☐ Conventional Machine Learning (CML)  ▨ Artificial Neural Network (ANN) & Deep Learning (DL)

**Figure 11:** Technique Classification Matrix

### 3.1.3. Algorithm Selection

The Algorithm classification mechanism is designed to facilitate Algorithm selection by traversing Figure 9 from the top to the bottom as illustrated in Figure 12. Since this section is rooted in a) theory and not extracted data, b) is a component of the design of the ontology, and c) is separable and distinct from Solution formulation, it is presented with the ontology rather than with extracted data or in the Guidelines or Conclusions sections. There are two anticipated use cases for Algorithm selection: i) A practitioner attempting to resolve a new Problem by employing Solutions that were effective at resolving similar Problems, and ii) A practitioner attempting to resolve a Problem or improve an existing Solution by exploring Solutions which are outside the bounds of current knowledge.

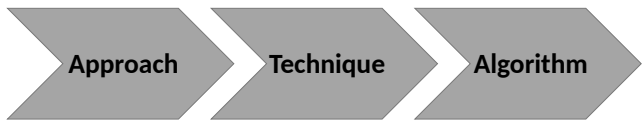

**Figure 12:** Algorithm Selection Process

When attempting to resolve a new Problem by employing Solutions that were effective at resolving similar Problems (i), efficient consumption of existing knowledge is required. The results presented in the Problem Domain and Solution Domain sections, along with figures in the Appendix, attempt to clearly convey actionable information regarding what is popular and has been attempted while specifying reference material where further reading can be pursued as listed in the Algorithms section. In this case, this selection process provides less utility since decisions are driven by existing knowledge.

When attempting to formulate a Solution or improve an existing Solution by exploring Solutions which are outside the bounds of current knowledge (ii), a procedure driving toward a desired outcome is preferred. In this case, the desired outcome is to strategically execute a process of elimination in such a way that the total amount of resources and time consumed to identify a Solution which resolves the Problem is decreased. The *first step* is to eliminate Approaches which are not compatible with the Problem or neighboring task constraints based on output dimensionality constraints and label requirements. The *second step* is to select a single Algorithm from each Technique in selected Approaches and compare their performance. This second step is based on the assumption that Algorithms which are in the same Technique will perform more similarly than algorithms in other Techniques since grouping within a Technique is based on similarity, and therefore that Algorithm's performance represents proxy performance for the entire Technique. The *final step* is to compare the performance of each Algorithm within the Technique of the best performing Algorithm from the second step, to determine the overall best performing Algorithm or identify which Algorithms contribute to a Solution which resolves the Problem.

The fragile assumption of common Technique performance is rooted in a tradeoff of scale and limited by subjective classification. By increasing the quantity of Techniques (and thereby making them finer-grained), the amount of trials required and therefore resources consumed to arrive at an acceptable Solution increase along with the probability of finding a better performing Algorithm. By decreasing the quantity of Techniques (and thereby making them coarser-grained), the resources consumed and probability of finding a better performing Algorithm decrease. The desired risk of overlooking a better performing Algorithm could be targeted dynamically by generating a new set of Techniques with the appropriate quantity of total Techniques, which thereby affects the grain size of Algorithms per Technique. Furthermore, this could be extended into a procedure of progressively increasing the quantity of Techniques in successive iterations until a Solution which resolves the Problem is found, further reducing the total number of trials. By extension, this could be combined with and expanded into a system of automated structure exploration where performance can be optimized through generative trial and error of different permutations of Algorithm and/or Solution structure.

However, it is important to identify that in manufacturing, the objective is likely to efficiently identify a Solution which is *good enough* to resolve the Problem, rather than identify the best performing Solution. Hence, a priori Technique quantity optimization's ability to provide a justifiable return on investment based on subjective classification is questionable. Furthermore, even though the granularity of Techniques can be adjusted to target an optimized outcome during the selection process, classification based on subjective notions of similarity carries the risk of grouping Algorithms which appear to be similar but in practice do not perform similarly and could lead to a lack of visibility of fruitful Algorithms. For this reason, developing objective metrics for similarity based on the structure of an Algorithm would enhance this method of Algorithm selection and decrease the risk of overlooking viable Algorithms which underpin Solutions that resolve the target Problem.

Considerations when applying this method during Solution formulation are expanded upon in the Guidelines section.

### 3.1.4. Use Case Example

To provide a concrete example of the proposed ontology, a surveyed paper is used to illustrate a use case. Li and Gu developed a smart fault-detection system for an industrial ball-bearing system [79]. The system was developed to enable the real-time detection and identification of bearing faults in rotating machinery so proactive action can be taken to prevent a reduction in profit driven by production losses in the form of wasted resources. The problem statement consisted of transforming real-time multivariate vibration sensor data into the detection of whether a fault has occurred, and if so, identifying the type of fault. In the Guidelines section, we recommend separating compound Problems like this into two separate and distinct Problems at a technical level of abstraction, such as i) Whether a fault has occurred, and ii) Identifying the type of fault, and not at a business level of abstraction such as profit optimization and waste reduction. The Problem and Solution domains are illustrated in Figure 13, and the anatomy of each task is illustrated in Figure 14. These figures are intended to illustrate a use case example and how a directed graph of tasks in a use case report might be illustrated in the future. In addition, these figures are not guaranteed to be accurate with respect to the literature from which they were derived.

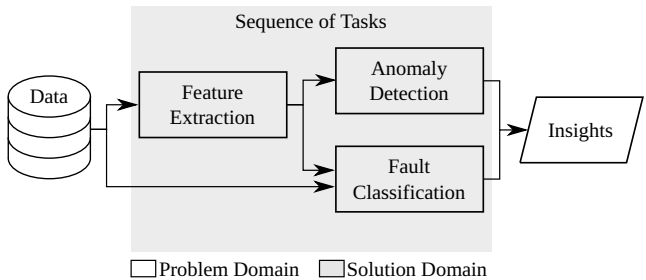

**Figure 13:** Use Case Tasks and Domains

The first task illustrated in Figure 14 focused on feature extraction used a Chen-Lee system to execute chaotic mapping which transformed the time-series vibration sensor data into abstract multi-dimensional features, followed by generating a 3D phase portrait of those features. Subsequently, the next task focused primarily on anomaly detection before calculating the average Euclidean distance of the phase portrait. The final task focused on fault classification using k-means to cluster based on chaotic mapping features and the phase portrait. By combining anomaly detection and fault classification task insights, the system was able to detect faults and successfully identify their type.

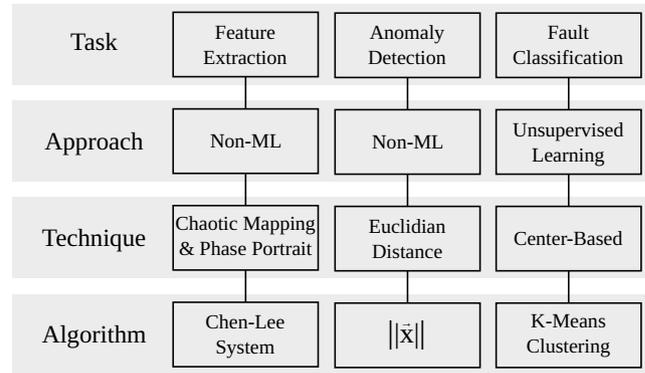

**Figure 14:** Use Case Task Structures

## 3.2. Problem Domain

During data extraction and analysis, a number of distinct themes emerged in the extracted data, and a number of challenges presented themselves with respect to conveying meaningful insights. To enable correlations and distinctions to be made effectively, we formulated classes for each attribute in order to highlight specific cross-sections of the data which were not easily discernible. In some cases, the formulation of attribute values was necessary for the same reasons. The formulation of attribute values and classes are representative of the extracted data and are not intended to be a component of the proposed ontology. We formulated the classes based on an analysis of the data itself, our classes are a result of and extension of our analysis, and our classes may not be transferable to or provide the same insights if this analysis is repeated on a larger scale in the future. While targeted figures are used to hone in on areas of interest in this section, comprehensive figures are included in the Appendix to ensure transparency surrounding the classification of attributes.

### 3.2.1. Problems

During data extraction, a number of distinct problems emerged. Due to the complexity of and tradeoffs in simplifying authors' research descriptions into problem statements to enable quantitative correlations and distinctions, we formulated problem attribute values and classes after the detailed review and data extraction steps. The formulation of these attribute values and classes were based on our best efforts to provide meaningful cross-sections of data, and the definitions used to discern between them are listed in Table 5. In concert, some problem classes consistently correlated with the use of specific jargon, which are listed in Table 6 and expanded upon in the Problem Jargon section.

| Class | Problem | Definition |
|---|---|---|
| Event & State Analysis | Diagnostics & Prognostics | Problems focusing on the identification of an event or state that has already occurred or a prediction of which event or state might occur. |
| | Change Detection | Problems focusing on identifying whether a state has changed or an event has occurred. |
| | Change Detection, Diagnostics, & Prognostics | Problems focused on identifying whether a state has changed or an event has occurred, then identifying the event or state change that has occurred or predicting a subsequent change that will occur. |
| Quality Control | Quality Control | Problems focused on quality control improvements. |
| Optimization | Process Optimization | Problems focused on optimizing a single manufacturing process. |
| | Factory/ Network Optimization | Problems focused on optimizing a collection or network of manufacturing processes. |

**Table 5:** Problem Classes

| Problem Class | Common Jargon |
|---|---|
| Diagnostics & Prognostics | Failure Prediction, Fault Diagnosis, Fault Classification, Tool Wear Classification, Condition Monitoring, Activity Recognition, Action Recognition, Object Identification |
| Change Detection | Change Detection, Anomaly Detection, Fault Detection, State Detection, Event Detection, Cycle Detection, Pattern Detection |

**Table 6:** Common Problem Jargon

Diagnostics & Prognostics was the most frequently encountered Problem, and Event & State Analysis was the most frequently encountered class, across both univariate and multivariate input dimensionalities, as shown in Figure 15. Quality Control was also of note, but primarily in the univariate space. A comprehensive matrix of Problems can be found in Figure 33 in the Appendix.

| | | | Input Class | | |
|---|---|---|---|---|---|
| | Class/Attribute | | Univariate | Multivariate | Total |
| Class | Event & State Analysis | | 40 | 40 | 80 |
| | Quality Control | | 16 | 3 | 19 |
| | Optimization | | 10 | 8 | 18 |
| | Total | | 66 | 51 | 117 |
| Problem Attribute | Diagnostics & Prognostics | | 23 | 17 | 40 |
| | Change Detection | | 16 | 14 | 30 |
| | Quality Control | | 16 | 3 | 19 |
| | Change Detection, Diagnostics, & Prognostics | | 1 | 9 | 10 |
| | Process Optimization | | 3 | 6 | 9 |
| | Factory/Network Optimization | | 7 | 2 | 9 |
| | Total | | 66 | 51 | 117 |

**Figure 15:** Problem Matrix

### 3.2.2. Industries

When analyzing the industries in which problems were addressed, we examined multiple existing industry categorization methods including those provided by the U.S. Department of Labor, NASA, and the Bureau of Labor Statistics. However, they did not yield meaningful results due to providing either too coarse or too fine of classes for this dataset. In addition, many of the Problems were applicable across a wide range of industries such as quality control and bearing health monitoring. To arrive at meaningful results, industries with industry-specific manufacturing problems were identified, while cross-applicable problems were lumped into a general manufacturing class, as illustrated in Figure 16. Strategically, this allows problems in general manufacturing and industry-specific manufacturing to be cross-sectioned individually. Both industry classes provided a relatively even distribution across input dimensionalities. However, while general manufacturing provided a relatively evenly distribution across problems, industry-specific classes are heavily weighted toward Event & State Analysis problems. A comprehensive matrix of industries can be found in Figure 32 in the Appendix.

| | | | Problem Class | | | | Input Class | | |
|---|---|---|---|---|---|---|---|---|---|
| | Class/Attribute | | Event & State Analysis | Quality Control | Optimization | Total | Univariate | Multivariate | Total |
| Class | General | | 31 | 19 | 14 | 64 | 40 | 24 | 64 |
| | Industry-Specific | | 49 | | 4 | 53 | 26 | 27 | 53 |
| | Total | | 80 | 19 | 18 | 117 | 66 | 51 | 117 |
| Industry Attribute | General Manufacturing | | 31 | 19 | 14 | 64 | 40 | 24 | 64 |
| | Automotive | | 9 | | | 9 | 7 | 2 | 9 |
| | Energy | | 4 | | 3 | 7 | 4 | 3 | 7 |
| | Oil and Gas | | 5 | | 1 | 6 | 3 | 3 | 6 |
| | Semiconductor | | 6 | | | 6 | 1 | 5 | 6 |
| | Healthcare | | 6 | | | 6 | 3 | 3 | 6 |
| | Agriculture and Food | | 5 | | | 5 | 1 | 4 | 5 |
| | Aerospace | | 5 | | | 5 | 2 | 3 | 5 |
| | Chemical | | 4 | | | 4 | 2 | 2 | 4 |
| | Construction & Utilities | | 3 | | | 3 | 1 | 2 | 3 |
| | Additive Manufacturing | | 2 | | | 2 | 2 | | 2 |
| | Total | | 80 | 19 | 18 | 117 | 66 | 51 | 117 |

**Figure 16:** Industry Matrix

### 3.2.3. Data Sources

When analyzing the data sources, both the combination of sources within a paper and individual sources across papers were analyzed. When extracting data from the 117 papers and considering multiple data sources per paper, a total of 150 instances of data sources were extracted. Data sources in combination are listed in Table 7, while the frequency of individual sources across all papers are listed in Figure 17.

When performing data extraction, a decision was made to curtail scope to focus on sensor type or the source of data if a specific sensor type was not identified (e.g. Statistical

Process Control (SPC)) based on the availability of detailed information, required labor to extract additional attributes, and perceived return on investment of specific attributes. With unlimited resources and comprehensive data, desired attributes would encompass the (i) phenomena, (ii) signal conditioning and processing, and (iii) acquisition parameters. (i) Phenomena would include the phenomena being studied, the instrument or sensor used to sense the phenomena, its construction/architecture, accuracy, precision, sensitivity, bandwidth (frequency range), and other relevant parameters based on the application. (ii) Signal conditioning and processing would include any analog signal conditioning or processing prior to Analog to Digital Conversion (ADC). (iii) Acquisition parameters would include the ADC type, reference type (e.g. single-ended, differential), sample rate, bit resolution, and other relevant parameters based on the application. These suggestions are extended in the Recommendations section. With the rising popularity of smart sensors and MicroElectroMechanical Systems (MEMS), the distinction between and identification of these three layers is less visible and discernible than traditional systems which utilize discrete components for each link in the signal chain. In the context of a MEMS which can incorporate the entire signal chain on a single Integrated Circuit (IC), the underlying methods employed might not be discernible and may even be closely held secrets.

| Data Sources | (Papers) |
|---|---|
| Vibration Sensor | 21 |
| SPC Data | 21 |
| Unspecified | 12 |
| Vision Sensor | 8 |
| Power Sensor | 7 |
| Temperature Sensor | 4 |
| Temperature Sensor, Pressure Sensor | 3 |
| Temperature Sensor, Pressure Sensor, Flow Rate Sensor | 2 |
| Flow Rate Sensor | 2 |
| Torque Sensor | 2 |
| Electroencephalogram (EEG) | 2 |
| Pressure Sensor | 2 |
| Radar | 2 |
| Motion Sensor | 2 |
| Electroencephalogram (EEG), Functional Near-Infrared Spectroscopy (fNIRS) | 1 |
| Vibration Sensor, Temperature Sensor | 1 |
| Power Sensor, Voltage Sensor, Current Sensor | 1 |
| Temperature Sensor, Voltage Sensor, Current Sensor | 1 |
| Global Positioning System (GPS) | 1 |
| SPC Data, Motion Sensor, Vibration Sensor | 1 |
| Global Positioning System (GPS), Motion Sensor, Optical Sensor | 1 |
| Vibration Sensor, Current Sensor | 1 |
| Power Sensor, Temperature Sensor | 1 |
| Pressure Sensor, Temperature Sensor, Flow Rate Sensor | 1 |
| Power Sensor, Temperature Sensor, Vibration Sensor | 1 |
| Pressure Sensor, Temperature Sensor, Volume Sensor | 1 |
| Magnetic Barkhausen Instrument | 1 |
| SPC Data, Force Sensor, Motion Sensor | 1 |
| Motion Sensor, Force Sensor | 1 |
| Current Sensor | 1 |
| Synthetic Data | 1 |
| Motion Sensor, Temperature Sensor, Humidity Sensor | 1 |
| Optical Sensor | 1 |
| Global Positioning System (GPS), Motion Sensor | 1 |
| Motion Sensor, Optical Sensor | 1 |
| Motion Sensor, Vibration Sensor | 1 |
| Electrocardiogram (ECG) | 1 |
| Quartz Crystal Microbalance (QCM) E-Nose | 1 |
| Strain Gauge | 1 |
| Volume Sensor | 1 |
| Force Sensor | 1 |
| **Total** | **117** |

**Table 7:** Data Source By Paper

As shown in Table 7 and Figure 17, vibration sensors and Statistical Process Control (SPC) data dominated as the most frequently examined data sources. However, while SPC data primarily correlated with univariate quality control problems in general manufacturing, kinetic data sources were used for Event & State Analysis across all industries and input dimensionalities. A comprehensive matrix of data sources can be found in Figure 34 in the Appendix.

| | Class/Attribute | Problem Class | | | | Industry Class | | | Input Class | | |
|---|---|---|---|---|---|---|---|---|---|---|---|
| | | Event & State Analysis | Optimization | Quality Control | Total | General | Industry-Specific | Total | Multivariate | Univariate | Total |
| Class | Kinetic | 35 | 7 | | 42 | 23 | 19 | 42 | 19 | 23 | 42 |
| | SPC | 3 | 3 | 17 | 23 | 21 | 2 | 23 | 5 | 18 | 23 |
| | Fluid | 15 | 2 | | 17 | 6 | 11 | 17 | 14 | 3 | 17 |
| | Electrical | 13 | 3 | 1 | 17 | 10 | 7 | 17 | 8 | 9 | 17 |
| | Temperature | 11 | 4 | 1 | 16 | 8 | 8 | 16 | 10 | 6 | 16 |
| | Unspecified | 7 | 5 | | 12 | 8 | 4 | 12 | 8 | 4 | 12 |
| | Vision | 9 | 2 | | 11 | 4 | 7 | 11 | 6 | 5 | 11 |
| | Medical | 6 | | | 6 | | 6 | 6 | 3 | 3 | 6 |
| | Position | 5 | | | 5 | | 5 | 5 | 3 | 2 | 5 |
| | Synthetic | 1 | | | 1 | 1 | | 1 | | 1 | 1 |
| | **Total** | **105** | **26** | **19** | **150** | **81** | **69** | **150** | **76** | **74** | **150** |
| Attribute | Vibration Sensor | 25 | 1 | | 26 | 15 | 11 | 26 | 10 | 16 | 26 |
| | SPC Data | 3 | 3 | 17 | 23 | 21 | 2 | 23 | 5 | 18 | 23 |
| | Temperature Sensor | 11 | 4 | 1 | 16 | 8 | 8 | 16 | 10 | 6 | 16 |
| | Unspecified | 7 | 5 | | 12 | 8 | 4 | 12 | 8 | 4 | 12 |
| | Power Sensor | 8 | 2 | | 10 | 5 | 5 | 10 | 5 | 5 | 10 |
| | Motion Sensor | 7 | 3 | | 10 | 6 | 4 | 10 | 6 | 4 | 10 |
| | Pressure Sensor | 8 | 1 | | 9 | 5 | 4 | 9 | 8 | 1 | 9 |
| | Vision Sensor | 7 | 1 | | 8 | 2 | 6 | 8 | 4 | 4 | 8 |
| | Flow Rate Sensor | 5 | | | 5 | | 5 | 5 | 3 | 2 | 5 |
| | Current Sensor | 3 | 1 | | 4 | 3 | 1 | 4 | 2 | 2 | 4 |
| | Optical Sensor | 2 | 1 | | 3 | 2 | 1 | 3 | 2 | 1 | 3 |
| | Global Positioning System (GPS) | 3 | | | 3 | | 3 | 3 | 1 | 2 | 3 |
| | Force Sensor | 1 | 2 | | 3 | 2 | 1 | 3 | 2 | 1 | 3 |
| | Electroencephalogram (EEG) | 3 | | | 3 | | 3 | 3 | 2 | 1 | 3 |
| | Voltage Sensor | 2 | | | 2 | 1 | 1 | 2 | 1 | 1 | 2 |
| | Torque Sensor | 1 | 1 | | 2 | | 2 | 2 | 1 | 1 | 2 |
| | Radar | 2 | | | 2 | | 2 | 2 | 2 | | 2 |
| | Volume Sensor | 2 | | | 2 | | 2 | 2 | 2 | | 2 |
| | Magnetic Barkhausen Instrument | | | 1 | 1 | 1 | | 1 | | 1 | 1 |
| | Humidity Sensor | | 1 | | 1 | 1 | | 1 | 1 | | 1 |
| | Synthetic Data | 1 | | | 1 | 1 | | 1 | | 1 | 1 |
| | Strain Gauge | 1 | | | 1 | | 1 | 1 | | 1 | 1 |
| | Electrocardiogram (ECG) | 1 | | | 1 | | 1 | 1 | | 1 | 1 |
| | Quartz Crystal Microbalance (QCM) E-Nose | 1 | | | 1 | | 1 | 1 | | 1 | 1 |
| | Functional Near-Infrared Spectroscopy (fNIRS) | 1 | | | 1 | | 1 | 1 | 1 | | 1 |
| | **Total** | **105** | **26** | **19** | **150** | **81** | **69** | **150** | **76** | **74** | **150** |

**Figure 17:** Data Source Matrix

### 3.2.4. Problem Jargon

A driving factor for the formulation of the ontology presented in the Ontology section was to provide a structure within which underlying concepts could be analyzed without the biases of terminology. The term Jargon is used to identify the terminology which may be inconsistently used and therefore be difficult for practitioners to understand. We selected a single Jargon for each paper based on our subjective notion of the defining terminology of that paper.

In this section, Jargon is correlated to Problem domain attributes to illustrate trends in the use of specific Jargon. As highlighted in Figure 18, Jargon such as Anomaly Detection and Fault Detection intuitively correlate with Change Detection across all industries. Likewise, Pattern Recognition, Time-Series Classification, and Fault Diagnosis intuitively correlate with Diagnostics & Prognostics across all industries. However, Binary Classification and Strip Breakage Detection were primarily found for Optimization Problems which do not provide a clear intuitive correlation, and may indicate a mismatch between the stated Problem which led to an optimization classification and the intended insights gained from the application of time-series analytics.

| Class/Attribute | Attribute | Problem Attribute | | | | | | Total | Industry Class | | Total |
|---|---|---|---|---|---|---|---|---|---|---|---|
| | | Diagnostics & Prognostics | Change Detection | Quality Control | Change Detection, Diagnostics, & Prognostics | Process Optimization | Factory/Network Optimization | | General | Industry-Specific | |
| Jargon Attribute | Control Chart Pattern Recognition | 1 | | 17 | | 1 | | 19 | 18 | 1 | 19 |
| | Anomaly Detection | 1 | 15 | | | | 1 | 17 | 8 | 9 | 17 |
| | Fault Detection | 2 | 6 | | 4 | 1 | | 13 | 7 | 6 | 13 |
| | Pattern Recognition | 8 | 1 | 1 | | 1 | 2 | 13 | 7 | 6 | 13 |
| | Time Series Classification | 5 | | | | 3 | 2 | 10 | 5 | 5 | 10 |
| | Fault Diagnosis | 6 | | | | | | 6 | 4 | 2 | 6 |
| | Fault Detection and Diagnosis | 1 | | | 2 | 1 | | 4 | 1 | 3 | 4 |
| | Time Series Clustering | 1 | | | | | 2 | 3 | | 3 | 3 |
| | Binary Classification | 1 | | | | 1 | | 2 | 1 | 1 | 2 |
| | Fault Detection and Classification (FDC) | 2 | | | | | | 2 | | 2 | 2 |
| | Damage Detection | | 1 | | | 1 | | 2 | 1 | 1 | 2 |
| | Change Detection | | 1 | 1 | | | | 2 | 1 | 1 | 2 |
| | Machine State Detection | | 2 | | | | | 2 | 2 | | 2 |
| | Human Activity Recognition (HAR) | 1 | | | | | | 1 | | 1 | 1 |
| | Earthquake Detection | 1 | | | | | | 1 | | 1 | 1 |
| | Annode Effect (AE) Detection | 1 | | | | | | 1 | | 1 | 1 |
| | Strip Breakage Detection | | | | 1 | | | 1 | 1 | | 1 |
| | Tool Wear Detection | 1 | | | | | | 1 | 1 | | 1 |
| | Unbalance Detection | | 1 | | | | | 1 | 1 | | 1 |
| | Disease Detection | 1 | | | | | | 1 | | 1 | 1 |
| | Fault Diagnosis and Prognosis | | | | | 1 | | 1 | 1 | | 1 |
| | Semantic Segmentation | 1 | | | | | | 1 | | 1 | 1 |
| | Break Detection | | 1 | | | | | 1 | 1 | | 1 |
| | Target Recognition | 1 | | | | | | 1 | | 1 | 1 |
| | Night Setback Detection | | | | | | 1 | 1 | 1 | | 1 |
| | Activity Identification | 1 | | | | | | 1 | | 1 | 1 |
| | Fault Detection and Isolation (FDI) | | | | 1 | | | 1 | 1 | | 1 |
| | Event Detection | | 1 | | | | | 1 | 1 | | 1 |
| | Activity Recognition | 1 | | | | | | 1 | | 1 | 1 |
| | Signal Classification | 1 | | | | | | 1 | | 1 | 1 |
| | Kick Detection | | | | 1 | | | 1 | 1 | | 1 |
| | Fault Detection and Identification | | | | 1 | | | 1 | 1 | | 1 |
| | Process Pattern Recognition | | 1 | | | | | 1 | 1 | | 1 |
| | **Total** | 40 | 30 | 19 | 10 | 9 | 9 | 117 | 64 | 53 | 117 |

**Figure 18:** Problem Jargon Matrix

While there is a strong correlation between the terminology used and the associated problem, the outliers highlight the inconsistency of subjective terminology. For instance, while "Fault Detection" is primarily correlated to Change Detection Problems, it is also less frequently correlated to Diagnostics & Prognostics and Process Optimization Problems. This raises further questions regarding the accuracy of use case report problem statements, how they correlated to the actual problem being resolved, and how use case reports are interpreted. To overcome these challenges and enable deeper analysis, a more formal and objective approach to defining and documenting problem domain attributes is recommended. A comprehensive matrix of problem Jargon can be found in Figure 35 in the Appendix.

## 3.3. Solution Domain

This section presents results pertaining to the Solution domain of the proposed Ontology, following the procedure and structure of the Algorithm Selection section to facilitate the guidance provided in the Guidelines section.

### 3.3.1. Approaches

The Algorithm Classification section defines the Approach classification mechanism based on objective attributes and is illustrated in Figure 10. The results of Approach classification on extracted data are illustrated in Figure 19. When extracting data from the 117 papers and considering multiple Algorithms per paper, a total of 229 instances of Algorithms were extracted. In total, there were 76 unique Algorithms across all 117 papers. While multiple Algorithms per paper were considered and some papers designated a single Solution as the best performing, a quantitative understanding of relative performance was typically not provided, nor were results readily correlatable between papers. Due to the perceived limited utility and return on investment of performance analysis based on partial data, we chose to limit the scope of Solution domain analysis to popularity based on frequency of use to provide correlation between attributes. As Figure 19 indicates, SL Approaches receive significantly more attention than UL Approaches among the reviewed papers. Reasons for this may be driven by the explainability and usefulness of SL Approaches, and the abundance of labeled data in manufacturing in comparison to other fields which is generated by readily available subject matter experts [86].

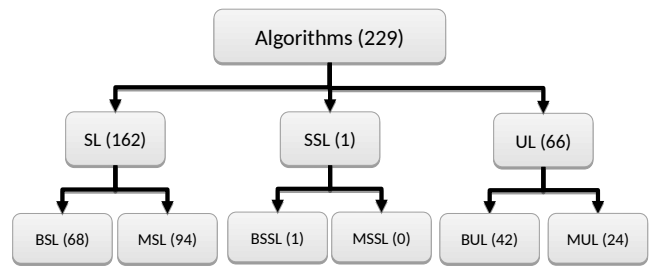

**Figure 19:** Approach Distribution

As discussed in the Algorithm Classification section, ANN and DL techniques are of special interest. To provide a cross-section of this attribute, techniques are classified based upon whether they use conventional or ANN/DL methods, as illustrated in Figure 20.

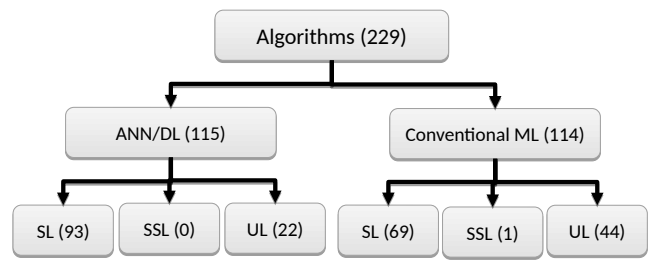

**Figure 20:** Technique Class Approach Distribution

Time-Series Pattern Recognition in Smart Manufacturing Systems: A Literature Review and Ontology*3.3.2. Techniques*

When formulating Techniques, we considered multiple existing classifications. The formulated Techniques are not intended to indicate a preference of one set of classes over another, but to illustrate the concept of Techniques and provide a fixed set of Techniques for result reference material generation to facilitate Algorithm selection. Until an objective method for Technique classification is developed and while research in this field continues, we expect subjective sets of Techniques to continue to grow and evolve.

To understand the popularity of Techniques and their correlation with Approaches, Figure 21 provides a cross-section exposing Feed Forward Neural Networks (FFNN), Convolutional Neural Networks (CNN), and Recurrent Neural Networks (RNN) as the most frequently explored Techniques. In addition, Figure 21 illustrates the even distribution of ANN/DL Techniques and Conventional ML Techniques, as well as the domination of SL approaches over UL and SSL approaches. In tandem, Techniques appeared evenly distributed with respect to output dimensionality.

| | Class/Attribute | Approach | | | | | | | | | | | |
|---|---|---|---|---|---|---|---|---|---|---|---|---|---|
| | | Class | | | | Class | | | Attribute | | | | |
| | | SL | UL | SSL | Total | M | B | Total | MSL | BSL | BUL | MUL | BSSL | Total |
| Class | ANN/DL | 93 | 22 | | 115 | 60 | 55 | 115 | 59 | 34 | 21 | 1 | | 115 |
| | Conventional | 69 | 44 | 1 | 114 | 58 | 56 | 114 | 35 | 34 | 21 | 23 | 1 | 114 |
| | Total | 162 | 66 | 1 | 229 | 118 | 111 | 229 | 94 | 68 | 42 | 24 | 1 | 229 |
| Technique Attribute | FF Neural Networks | 33 | 3 | | 36 | 25 | 11 | 36 | 25 | 8 | 3 | | | 36 |
| | Convolutional Neural Networks (CNN) | 29 | 4 | | 33 | 19 | 14 | 33 | 19 | 10 | 4 | | | 33 |
| | Recurrent Neural Networks (RNN) | 27 | 3 | | 30 | 14 | 16 | 30 | 14 | 13 | 3 | | | 30 |
| | Decision Tree-Based | 29 | | | 29 | 21 | 8 | 29 | 21 | 8 | | | | 29 |
| | Instance-Based | 28 | | | 28 | 12 | 16 | 28 | 12 | 16 | | | | 28 |
| | Center-Based | | 16 | | 16 | 12 | 4 | 16 | | | 4 | 12 | | 16 |
| | Encoder/Decoder Based | 4 | 8 | | 12 | 1 | 11 | 12 | 1 | 3 | 8 | | | 12 |
| | Statistical | 8 | | | 8 | 2 | 6 | 8 | 2 | 6 | | | | 8 |
| | Connectivity-Based | | 8 | | 8 | 6 | 2 | 8 | | | 2 | 6 | | 8 |
| | Density-Based | | 7 | | 7 | 3 | 4 | 7 | | | 4 | 3 | | 7 |
| | Distribution-Based | | 5 | | 5 | 1 | 4 | 5 | | | 4 | 1 | | 5 |
| | Regression | 4 | | | 4 | | 4 | 4 | | 4 | | | | 4 |
| | Dimension Reduction | | 4 | | 4 | 1 | 3 | 4 | | | 3 | 1 | | 4 |
| | Constraint-Based | | 4 | | 4 | | 4 | 4 | | | 4 | | | 4 |
| | Generative Adversarial Models | | 4 | | 4 | 1 | 3 | 4 | | | 3 | 1 | | 4 |
| | Semi Supervised Detection of Outliers (SSDO) | | | 1 | 1 | 1 | | 1 | | | | | 1 | 1 |
| | Total | 162 | 66 | 1 | 229 | 118 | 111 | 229 | 94 | 68 | 42 | 24 | 1 | 229 |

**Figure 21:** Technique Results

Our review shows a higher preference toward ANN/DL Techniques compared to conventional ML Techniques among investigated papers. One reason behind this preference could be the concept of time-awareness. Conventional ML classification Algorithms such as Support Vector Machine (SVM) and Random Forest (RF) cannot comprehend the notion of time and the inherent interdependency between sequential data points. As a result, they require an additional step of feature extraction to overcome this issue. By contrast, ANN/DL Algorithms such as RNN, Gated Recurrent Unit (GRU), and Long Short-Term Memory (LSTM) are designed to handle structured data and are capable of overcoming this issue without additional feature extraction. As a result of hidden cells which can consider information from previous cells, they are considered to be time-aware algorithms and present a clear advantage in time-series analytics.

Another reason for the preference toward ANN/DL could be the need for feature extraction for conventional ML models. For conventional ML Algorithms, feature extraction is required before model training, while in DL, feature extraction and model training can be conducted concurrently. The feature extraction task in conventional ML algorithms is a time-consuming and expensive process requiring domain knowledge, hand-crafted features that may be specific to each application, and the risk of losing information during the process. This is especially true for large and complex datasets. Conversely, ANN/DL Algorithms are capable of performing feature extraction tasks automatically. These functions can be learned dynamically through the network rather than being individually assembled, giving better results with little domain knowledge. In summary, ANN/DL focuses on extracting only the relevant features, improving performance and accuracy [94].

However, despite the demonstrated performance of ANN/DL Algorithms and their improved accuracy and robustness in pattern recognition compared to conventional ML algorithms, they are still not readily applicable in all cases, in part due to their black-box nature. In other words, it is difficult to explain the rationale for the decisions made by deep neural networks, making it difficult to evaluate their reliability. Explainable Artificial Intelligence (XAI) is especially important in manufacturing and industrial settings and has attracted attention recently in the research community [40]. This is likely driven by concerns regarding safety, responsibility, and risk management of difficult to explain or unexplainable ANN/DL algorithms.

To better visualize which Techniques have been applied for a given Approach, Figure 22 delivers the same information in the format of Figure 11 from the Algorithm Classification section. By using these figures, a practitioner should be able to identify Techniques which are applicable based on chosen Approaches. A comprehensive matrix of Techniques can be found in Figure 20 in the Appendix.

| | | Learning Type | | |
|---|---|---|---|---|
| | | Supervised Learning (SL) | Semi-Supervised Learning (SSL) | Unsupervised Learning (UL) |
| Output Dimensionality | Binary | (34) Instance-Based (16) Decision Tree-Based (8) Statistical (6) Regression (4) | (1) Semi-Supervised Detection of Outliers (SSDO) (1) | (21) Constraint-Based (4) Density-Based (4) Center-Based (4) Distribution-Based (4) Dimension Reduction (3) Connectivity-Based (2) |
| | Binary | (34) Recurrent Neural Networks (RNN) (13) Convolutional Neural Networks (CNN) (10) Feed Forward Neural Networks (FFN) (8) Autoencoders (AE) (3) | (0) | (21) Autoencoders (AE) (8) Convolutional Neural Networks (CNN) (4) Recurrent Neural Networks (RNN) (3) Generative Adversarial Models (3) Feed Forward Neural Networks (FFN) (3) |
| | Multiary | (35) Decision Tree-Based (21) Instance-Based (12) Statistical (2) | (0) | (23) Center-Based (12) Connectivity-Based (6) Density-Based (3) Dimension Reduction (1) Distribution-Based (1) |
| | Multiary | (59) Feed Forward Neural Networks (FFN) (25) Convolutional Neural Networks (CNN) (19) Recurrent Neural Networks (RNN) (14) Autoencoders (AE) (1) | (0) | (1) Generative Adversarial Models (1) |
| | | Fully | Partially | None |
| | | Label Requirement | | |

□ Conventional Machine Learning (CML)   ▨ Artificial Neural Network (ANN) & Deep Learning (DL)

**Figure 22:** Technique Classification Results Matrix

Farahani et al.: *Preprint submitted to Elsevier*   Page 13 of 41

### 3.3.3. Algorithms

To understand the popularity of Algorithms and their correlation with Approaches, Figure 23 provides a cross-section exposing CNN, Multi-Layer Perceptrons (MLP), and Random Forest (RF) as the most frequently explored Algorithms. To facilitate Algorithm selection based on target Approaches, Table 8, Table 9, Table 10, Table 11, and Table 12 provide a cross-section of applicable Algorithms cross-sectioned by Technique. A comprehensive matrix of Algorithms can be found in Figure 20, and comprehensive matrices of Algorithms by Technique can be found in Table 13, Table 14, Table 15, Table 16, and Table 17 in the Appendix.

Figure 23: Algorithm Results

| Technique | Class | Algorithm |
|---|---|---|
| Instance-Based (16) | Conventional | K-Nearest Neighbor (KNN) (5) |
| | | Support Vector Machines (SVM) (5) |
| | | One-Class SVM (3) |
| | | Dynamic Time Warping (DTW) (1) |
| | | Support Vector Data Description (SVDD) (1) |
| | | Learning Vector Quantization (LVQ) (1) |
| Recurrent Neural Networks (RNN) (13) | ANN/DL | LSTM Network (7) |
| | | CNN-LSTM Network (2) |
| | | BDLSTM (Bi-LSTM) network (2) |
| | | Gated Recurrent Unit (GRU) (1) |
| | | RNN (1) |
| Convolutional Neural Networks (CNN) (10) | ANN/DL | Convolutional Neural Network (CNN) (6) |
| | | 1D-CNN (2) |
| | | Fully Convolutional Networks (FCN) (1) |
| | | Temporal Convolutional Networks (TCN) (1) |
| Decision Tree-Based (8) | Conventional | Random Forest (RF) (6) |
| | | eXtreme Gradient Boosting (XGBoost) (1) |
| | | Gradient Boosting Classifier (GBM) (1) |
| FF Neural Networks (8) | ANN/DL | Multi Layer Perceptron's (MLP) (4) |
| | | Fully Connected Neural Networks (FCNN) (2) |
| | | Back Propagation Network (BPN) (2) |
| Statistical (6) | Conventional | Linear Discriminant Analysis (LDA) (2) |
| | | Quadratic Discriminant Analysis (QDA) (2) |
| | | Naive Bayes Classifier (2) |
| Regression (4) | Conventional | Logistic Regression (LR) (4) |
| Encoder/Decoder Based (3) | ANN/DL | Variational AE (VAE) (1) |
| | | Deep Stacked Autoencoder (DSAE) (1) |
| | | Attention Mechanism (1) |

Table 8: BSL Algorithms By Technique

| Technique | Class | Algorithm |
|---|---|---|
| FF Neural Networks (25) | ANN/DL | Multi Layer Perceptron's (MLP) (14) |
| | | Back Propagation Network (BPN) (4) |
| | | Radial Basis Function Neural Networks (RBFNNs) (2) |
| | | Fully Connected Neural Networks (FCNN) (2) |
| | | Feedforward Fully Connected Spiking Neural Network (1) |
| | | ESSM (1) |
| | | Probabilistic Neural Networks (PNNs) (1) |
| Decision Tree-Based (21) | Conventional | Random Forest (RF) (11) |
| | | eXtreme Gradient Boosting (XGBoost) (3) |
| | | Gradient Boosting Classifier (GBM) (3) |
| | | Decision Tree (Dtree) (1) |
| | | Bagging Classifier (BAG) (1) |
| | | RUS Boosted Decision Trees (1) |
| | | Extreme Random Forest (ERF) (1) |
| Convolutional Neural Networks (CNN) (19) | ANN/DL | Convolutional Neural Network (CNN) (11) |
| | | 1D-CNN (3) |
| | | Residual Neural Network (ResNet) (2) |
| | | HHO-ConvNet (1) |
| | | FDC-CNN Model (1) |
| | | Explainable DCNN (1) |
| Recurrent Neural Networks (RNN) (14) | ANN/DL | LSTM Network (5) |
| | | CNN-LSTM Network (2) |
| | | Gated Recurrent Unit (GRU) (2) |
| | | BDLSTM (Bi-LSTM) network (2) |
| | | NARX Neural Network (1) |
| | | AtLSTMs (1) |
| | | Stacked LSTM (1) |
| Instance-Based (12) | Conventional | Support Vector Machines (SVM) (6) |
| | | K-Nearest Neighbor (KNN) (4) |
| | | Self-Organizing Map (SOM) (1) |
| | | Learning Vector Quantization (LVQ) (1) |
| Statistical (2) | Conventional | Naive Bayes Classifier |
| Encoder/Decoder Based (1) | ANN/DL | Stacked Denoising Autoencoder (SDAE) (1) |

Table 9: MSL Algorithms By Technique

| Technique | Class | Algorithm |
|---|---|---|
| Semi Supervised Detection of Outliers (SSDO) (1) | Conventional | COP K-Means Constrained Clustering Algorithm (1) |

Table 10: BSSL Algorithms By Technique

| Technique | Class | Algorithm |
|---|---|---|
| Encoder/Decoder Based (8) | ANN/DL | Encoder Decoder (E/D) (2) |
| | | RNN E/D (2) |
| | | Adversarial Autoencoder (AAE) (1) |
| | | Transformer Encoder (1) |
| | | ECHAD (Embedding-based CHAnge Detection) (1) |
| | | Variational Bi-LSTM AE (1) |
| Constraint-Based (4) | Conventional | Isolation Forest (iForest) (4) |
| Density-Based (4) | Conventional | Local Outlier Factor (LOF) (3) |
| | | Kernel Spectral Clustering (KSC) (1) |
| Center-Based (4) | Conventional | K-Means Clustering (3) |
| | | Histogram-Based Outlier Score (HBOS) (1) |
| Distribution-Based (4) | Conventional | Gaussian Mixture Models (GMM) (3) |
| | | Parzen Window (1) |
| Convolutional Neural Networks (CNN) (4) | ANN/DL | 1D-CNN (2) |
| | | Convolutional Neural Network (CNN) (2) |
| Dimension Reduction (3) | Conventional | Principal Component Analysis (PCA) (2) |
| | | Kernel PCA (KPCA) (1) |
| FF Neural Networks (3) | ANN/DL | Fully Connected Neural Networks (FCNN) (2) |
| | | Bayesian Interpretation of Neural Networks (BINN) (1) |
| Generative Adversarial Models (3) | ANN/DL | Generative Adversarial Network (GAN) (2) |
| | | Mode Seeking Generative Adversial Network (MSGAN) (1) |
| Recurrent Neural Networks (RNN) (3) | ANN/DL | RNN (1) |
| | | Stacked LSTM (1) |
| | | CNN-LSTM Network (1) |
| Connectivity-Based (2) | Conventional | Hierarchical TICC (1) |
| | | Hierarchical Clustering (1) |

**Table 11:** BUL Algorithms By Technique

| Technique | Class | Algorithm |
|---|---|---|
| Center-Based (12) | Conventional | K-Means Clustering (9) |
| | | K-MICA clustering (1) |
| | | Kernel K-means (1) |
| | | K-Medoids Clustering (1) |
| Connectivity-Based (6) | Conventional | Agglomerative Hierarchical Clustering (AGNES) (2) |
| | | Hierarchical TICC (2) |
| | | Weighted Pair Group Method with Arithmetic Mean (WPGMA) (1) |
| | | Spectral Clustering (1) |
| Density-Based (3) | Conventional | MDL clustering (1) |
| | | OPTICS (1) |
| | | DBSCAN (1) |
| Dimension Reduction (1) | Conventional | Self Organizing Map (SOM) (1) |
| Distribution-Based (1) | Conventional | Gaussian Mixture Models (GMM) (1) |
| Generative Adversarial Models (1) | ANN/DL | Generative Adversarial Network (GAN) (1) |

**Table 12:** MUL Algorithms By Technique

### 3.3.4. Solution Jargon

Similar to the Problem Jargon section, this section provides a correlation of Jargon, but with respect to Solution domain attributes. When evaluating papers, some Jargon displayed a strong correlation with Approach as illustrated in Figure 25. For instance, Anomaly Detection commonly refers to identifying rare values or sequences that deviate from expected or normal operating conditions without having labels or supervisory data to validate against, which strongly correlates with BUL [129]. By applying BUL, a cluster can be developed representing normal operating conditions, and performance lying outside that cluster can be identified as anomalous behavior. Similarly, Fault Detection commonly refers to identifying values or sequences that deviate from expected or normal operating conditions, but labels or supervisory data is available to validate against favoring a BSL Approach. This implies that the rarity of the fault, represented by how imbalanced the data is, likely has less of an impact on Algorithm performance since labels or supervisory data is available to validate against. Furthermore, the distinction between whether the term Anomaly Detection or Fault Classification is used is highly correlated to data imbalance. While Fault Detection was used to describe relatively frequent deviations from normal operating conditions which appear relatively frequently in the data, Anomaly Detection was used to describe rare deviations from normal operating conditions which rarely appear in the data.

In addition and focusing on multiary output dimensionalities, Machine State Detection strongly correlated with MUL, while Pattern Recognition and Time-Series Classification strongly correlated with MSL. During data extraction, common terminology associated with each Approach were identified, as shown in Figure 24.

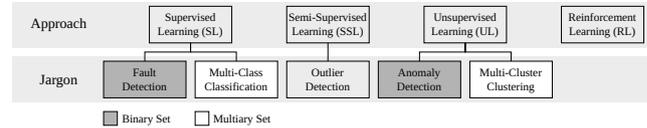

**Figure 24:** Common Jargon

While these strong correlations provide insight into how the terminology used relates to the implemented Solution, the outliers highlight the inconsistency of subjective terminology and the need for an objective classification mechanism to clearly define and illustrate underlying concepts. The Approach classification mechanism proposed in the Algorithm Classification section is intended to provide this functionality and enable decoupling from subjective terminology bias.

A similar challenge was encountered with terminology related to Algorithms. In a number of instances, different authors used different terminology to name Algorithms which appear structurally the same or extremely similar. At present, manufacturing-focused practitioners are reliant on computer science subject matter experts to provide nuanced and distilled analyses and comparisons of complex algorithms. To avoid this dependency, simplify application, and enable independent analysis of Algorithms, an objective method for analyzing the similarity of algorithms is necessary. Furthermore, decoupling the terminology of Algorithm names from objective differences such as structure to avoid subjective terminology bias could also be beneficial similar to our recommendation using Approaches in place of Jargon.

A comprehensive matrix of solution domain Jargon can be found in Figure 21 in the Appendix.

| Class/Attribute | | Approach Attribute | | | | | | Technique Class | | |
|---|---|---|---|---|---|---|---|---|---|---|
| | | Multiary Supervised Learning (MSL) | Binary Supervised Learning (BSL) | Binary Unsupervised Learning (BUL) | Multiary Unsupervised Learning (MUL) | Binary Semi-Supervised Learning (BSSL) | Total | ANN/DL | Conventional | Total |
| Jargon Attribute | Anomaly Detection | | 2 | 34 | | 1 | 37 | 20 | 17 | 37 |
| | Control Chart Pattern Recognition | 23 | | | 3 | | 26 | 21 | 5 | 26 |
| | Fault Detection | 5 | 14 | 3 | 2 | | 24 | 10 | 14 | 24 |
| | Time Series Classification | 15 | 7 | 1 | | | 23 | 10 | 13 | 23 |
| | Pattern Recognition | 10 | 5 | 1 | 4 | | 20 | 6 | 14 | 20 |
| | Machine State Detection | 2 | | | 10 | | 12 | 2 | 10 | 12 |
| | Fault Diagnosis | 8 | 2 | 1 | | | 11 | 5 | 6 | 11 |
| | Activity Recognition | 7 | | | | | 7 | 1 | 6 | 7 |
| | Fault Detection and Classification (FDC) | 1 | 4 | | | | 5 | 1 | 4 | 5 |
| | Fault Detection and Diagnosis | 1 | 3 | | 1 | | 5 | 4 | 1 | 5 |
| | Activity Identification | 4 | | | 1 | | 5 | 5 | | 5 |
| | Annode Effect (AE) Detection | | 5 | | | | 5 | | 5 | 5 |
| | Target Recognition | 4 | | | | | 4 | 4 | | 4 |
| | Human Activity Recognition (HAR) | 4 | | | | | 4 | 4 | | 4 |
| | Binary Classification | | 4 | | | | 4 | 2 | 2 | 4 |
| | Unbalance Detection | | 4 | | | | 4 | 2 | 2 | 4 |
| | Disease Detection | 3 | | | | | 3 | 2 | 1 | 3 |
| | Time Series Clustering | | | 3 | | | 3 | | 3 | 3 |
| | Damage Detection | | 3 | | | | 3 | 3 | | 3 |
| | Fault Detection and Identification | | 2 | 1 | | | 3 | | 3 | 3 |
| | Change Detection | 1 | | 1 | | | 2 | 2 | | 2 |
| | Break Detection | | 2 | | | | 2 | 2 | | 2 |
| | Fault Diagnosis and Prognosis | 2 | | | | | 2 | | 2 | 2 |
| | Strip Breakage Detection | | 2 | | | | 2 | 2 | | 2 |
| | Event Detection | | 2 | | | | 2 | 2 | | 2 |
| | Night Setback Detection | | 2 | | | | 2 | 2 | | 2 |
| | Process Pattern Recognition | 1 | | | | | 1 | 1 | | 1 |
| | Signal Classification | | 1 | | | | 1 | 1 | | 1 |
| | Earthquake Detection | | 1 | | | | 1 | | 1 | 1 |
| | Action Recognition | 1 | | | | | 1 | 1 | | 1 |
| | Fault Detection and Isolation (FDI) | 1 | | | | | 1 | 1 | | 1 |
| | Fault Classification | 1 | | | | | 1 | 1 | | 1 |
| | Tool Wear Detection | | 1 | | | | 1 | 1 | | 1 |
| | **Total** | 94 | 68 | 42 | 24 | 1 | 229 | 115 | 114 | 229 |

**Figure 25:** Solution Jargon Matrix

| | Class/Attribute | Approach Class | | | | Class | | | Technique Class | | |
|---|---|---|---|---|---|---|---|---|---|---|---|
| | | Supervised (SL) | Unsupervised (UL) | Semi-Supervised (SSL) | Total | Multiary (M) | Binary (B) | Total | ANN/DL | Conventional | Total |
| Problem Class | Event & State Analysis | 109 | 56 | 1 | 166 | 70 | 96 | 166 | 77 | 89 | 166 |
| | Optimization | 32 | 4 | | 36 | 22 | 14 | 36 | 18 | 18 | 36 |
| | Quality Control | 21 | 6 | | 27 | 26 | 1 | 27 | 20 | 7 | 27 |
| | **Total** | 162 | 66 | 1 | 229 | 118 | 111 | 229 | 115 | 114 | 229 |
| Problem Attribute | Diagnostics & Prognostics | 73 | 5 | | 78 | 51 | 27 | 78 | 34 | 44 | 78 |
| | Change Detection | 22 | 47 | 1 | 70 | 15 | 55 | 70 | 38 | 32 | 70 |
| | Quality Control | 21 | 6 | | 27 | 26 | 1 | 27 | 20 | 7 | 27 |
| | Process Optimization | 20 | 2 | | 22 | 13 | 9 | 22 | 12 | 10 | 22 |
| | Change Detection, Diagnostics, & Prognostics | 14 | 4 | | 18 | 4 | 14 | 18 | 5 | 13 | 18 |
| | Factory/Network Optimization | 12 | 2 | | 14 | 9 | 5 | 14 | 6 | 8 | 14 |
| | **Total** | 162 | 66 | 1 | 229 | 118 | 111 | 229 | 115 | 114 | 229 |
| Industry Class | General | 88 | 38 | | 126 | 77 | 49 | 126 | 65 | 61 | 126 |
| | Industry-Specific | 74 | 28 | 1 | 103 | 41 | 62 | 103 | 50 | 53 | 103 |
| | **Total** | 162 | 66 | 1 | 229 | 118 | 111 | 229 | 115 | 114 | 229 |
| Industry Attribute | General Manufacturing | 88 | 38 | | 126 | 77 | 49 | 126 | 65 | 61 | 126 |
| | Automotive | 9 | 8 | | 17 | 2 | 15 | 17 | 8 | 9 | 17 |
| | Oil and Gas | 14 | 1 | | 15 | 4 | 11 | 15 | 6 | 9 | 15 |
| | Healthcare | 13 | 1 | | 14 | 11 | 3 | 14 | 8 | 6 | 14 |
| | Semiconductor | 10 | 2 | | 12 | 5 | 7 | 12 | 5 | 7 | 12 |
| | Chemical | 8 | 2 | | 10 | 2 | 8 | 10 | 3 | 7 | 10 |
| | Energy | 5 | 4 | | 9 | 4 | 5 | 9 | 5 | 4 | 9 |
| | Aerospace | 7 | 2 | | 9 | 8 | 1 | 9 | 8 | 1 | 9 |
| | Construction & Utilities | 1 | 6 | 1 | 8 | | 8 | 8 | 3 | 5 | 8 |
| | Agriculture and Food | 5 | 2 | | 7 | 5 | 2 | 7 | 3 | 4 | 7 |
| | Additive Manufacturing | 2 | | | 2 | 2 | | 2 | 1 | 1 | 2 |
| | **Total** | 162 | 66 | 1 | 229 | 118 | 111 | 229 | 115 | 114 | 229 |
| Input Class | Univariate | 86 | 31 | 1 | 118 | 65 | 53 | 118 | 62 | 56 | 118 |
| | Multivariate | 76 | 35 | | 111 | 53 | 58 | 111 | 53 | 58 | 111 |
| | **Total** | 162 | 66 | 1 | 229 | 118 | 111 | 229 | 115 | 114 | 229 |

**Figure 26:** Problem-Solution Matrix

### 3.4. Problem-Solution Relationship

To understand the relationship between the Problem and Solution domains, attributes from each domain were analyzed as shown in Figure 26 and Figure 27. When extracting data from the 117 papers from the Problem domain and 229 instances of Algorithms from the Solution domain, and considering multiple data sources per paper, a total of 308 instances of data sources were extracted.

As illustrated in Figure 26, Change Detection was heavily correlated to UL, whereas all other Problems were dominated by SL. This may be driven by uncertainty about a subject matter expert's ability to envision and capture all potential states and changes that could occur in a system through labeling, or a general openness to the existence of unknown states and events.

Comprehensive matrices can be found in the Appendix illustrating domain borders in Figure 19, and Solution data sources can be found in Figure 18.

|  |  | Approach | | | | | | | Technique | | |
|---|---|---|---|---|---|---|---|---|---|---|---|
|  | Class/Attribute | Class | | | | Class | | | Class | | |
|  |  | Supervised (SL) | Unsupervised (UL) | Semi-Supervised (SSL) | Total | Binary (B) | Multiary (M) | Total | Conventional | ANN/DL | Total |
| **Class** | Kinetic | 63 | 14 |  | 77 | 43 | 34 | 77 | 27 | 50 | 77 |
|  | Fluid | 22 | 18 | 1 | 41 | 36 | 5 | 41 | 24 | 17 | 41 |
|  | Temperature | 26 | 12 |  | 38 | 29 | 9 | 38 | 31 | 7 | 38 |
|  | Electrical | 24 | 9 |  | 33 | 20 | 13 | 33 | 21 | 12 | 33 |
|  | Unspecified | 21 | 12 |  | 33 | 12 | 21 | 33 | 22 | 11 | 33 |
|  | SPC | 28 | 4 |  | 32 | 5 | 27 | 32 | 6 | 26 | 32 |
|  | Vision | 25 | 5 |  | 30 | 9 | 21 | 30 | 19 | 11 | 30 |
|  | Position | 11 | 1 |  | 12 | 4 | 8 | 12 |  | 12 | 12 |
|  | Medical | 10 | 1 |  | 11 | 4 | 7 | 11 | 8 | 3 | 11 |
|  | Synthetic |  |  | 1 | 1 |  | 1 | 1 | 1 |  | 1 |
|  | Total | 230 | 77 | 1 | 308 | 162 | 146 | 308 | 159 | 149 | 308 |
| **Data Source** | Vibration Sensor | 32 | 10 |  | 42 | 21 | 21 | 42 | 19 | 23 | 42 |
|  | Temperature Sensor | 26 | 12 |  | 38 | 29 | 9 | 38 | 31 | 7 | 38 |
|  | Unspecified | 21 | 12 |  | 33 | 12 | 21 | 33 | 22 | 11 | 33 |
|  | SPC Data | 28 | 4 |  | 32 | 5 | 27 | 32 | 6 | 26 | 32 |
|  | Motion Sensor | 24 |  |  | 24 | 12 | 12 | 24 | 6 | 18 | 24 |
|  | Pressure Sensor | 12 | 12 |  | 24 | 20 | 4 | 24 | 15 | 9 | 24 |
|  | Vision Sensor | 13 | 5 |  | 18 | 8 | 10 | 18 | 12 | 6 | 18 |
|  | Power Sensor | 8 | 6 |  | 14 | 8 | 6 | 14 | 8 | 6 | 14 |
|  | Optical Sensor | 12 |  |  | 12 | 1 | 11 | 12 | 7 | 5 | 12 |
|  | Current Sensor | 10 | 2 |  | 12 | 6 | 6 | 12 | 8 | 4 | 12 |
|  | Flow Rate Sensor | 8 |  | 1 | 9 | 8 | 1 | 9 | 6 | 3 | 9 |
|  | Volume Sensor | 1 | 6 |  | 7 | 7 |  | 7 | 3 | 4 | 7 |
|  | Global Positioning System (GPS) | 6 | 1 |  | 7 | 3 | 4 | 7 |  | 7 | 7 |
|  | Electroencephalogram (EEG) | 6 | 1 |  | 7 | 1 | 6 | 7 | 6 | 1 | 7 |
|  | Voltage Sensor | 5 | 1 |  | 6 | 6 |  | 6 | 5 | 1 | 6 |
|  | Force Sensor | 5 |  |  | 5 | 5 |  | 5 | 1 | 4 | 5 |
|  | Torque Sensor | 1 | 4 |  | 5 | 4 | 1 | 5 | 1 | 4 | 5 |
|  | Radar | 5 |  |  | 5 | 1 | 4 | 5 |  | 5 | 5 |
|  | Electrocardiogram (ECG) | 2 |  |  | 2 | 2 |  | 2 |  | 2 | 2 |
|  | Quartz Crystal Microbalance (QCM) E-Nose | 1 |  |  | 1 |  | 1 | 1 | 1 |  | 1 |
|  | Synthetic Data |  |  | 1 | 1 |  | 1 | 1 | 1 |  | 1 |
|  | Functional Near-Infrared Spectroscopy (fNIRS) | 1 |  |  | 1 | 1 |  | 1 | 1 |  | 1 |
|  | Strain Gauge | 1 |  |  | 1 | 1 |  | 1 |  | 1 | 1 |
|  | Magnetic Barkhausen Instrument | 1 |  |  | 1 |  | 1 | 1 | 1 |  | 1 |
|  | Total | 230 | 77 | 1 | 308 | 162 | 146 | 308 | 159 | 149 | 308 |

**Figure 27:** Data Source Problem-Solution Matrix

## 4. Guidelines

In this section, we extend our discussion and transmute key results and insights into simple and actionable guidelines. While the guidelines are high level, concise considerations for application of time-series pattern recognition in manufacturing are provided through a novel four step process.

### 4.1. Step 1: Survey Ontology Boundaries

As illustrated in Figure 6, data acquisition and decision making exist outside the Problem and Solution domains of the proposed ontology. To benefit from the presented ontology, the boundaries of said ontology must be assessed.

#### 4.1.1. Data Acquisition

Data acquisition for the Problem domain can acquire both historical and real-time data from a variety of sources, each with different constraints. Historical data can be downloaded from publically available repositories or be generated and retained privately. Although the public sharing of datasets is commonplace and can be a requirement for publication in the computer science field, a manufacturing-specific challenge is a wide-spread, strong hesitancy to share data [141]. The reasons are manifold and include cybersecurity, competitive landscape, prioritization, and subject knowledge.

There are ample amounts of publicly available time-series datasets covering a vast range of applications outside of manufacturing such as in the medical domain (e.g., ECG, heartbeat), the financial sector (e.g., exchange rates and stock trade), human activity recognition, and speech recognition applications. The University of California, Riverside (UCR) time-series archive [21], The University of East Anglia (UEA) multivariate time-series classification archive [7], and the UCR anomaly archive [150] are examples of these publicly available datasets. Each of these sources contains a vast quantity of datasets that can be used for research purposes. However, the number of manufacturing-related public datasets that can be used in time-series pattern recognition are limited and cover a narrow range of applications. In [106], five datasets were introduced that can be used for bearing fault detection and diagnosis applications. The Prognostics and Health Management (PHM) Society is another resource that hosts a data-driven competition every year and has a few public datasets that can be used for time-series pattern recognition research [48]. Due to the reluctance to share data and scarcity of publicly available datasets, manufacturing is missing out on valuable advantages afforded to the computer science field by strong data availability to advance the field as a whole.

One such advantage is in education, where public datasets enable students to gain experience, exposure, and proficiency working with real data. A lack of knowhow, data, and experienced talent to work with AI have been identified as critical challenges facing AI adoption, and the availability of public manufacturing datasets may be a key factor for overcoming those challenges through education [141][34]. An additional advantage is in benchmarking, where apples-to-apples, correlatable performance comparisons of algorithms can be executed using specified metrics and publicly available datasets. Without shared datasets and common metrics, an objective and thorough comparison of performance is unlikely. A final advantage is transparency through the ability to validate and verify other people's work. Within the community of parties who are in favor of sharing data and rigorous validation of methods, the exclusion of non-transparent methods may be viewed as a feature rather than a bug.

While the analysis of historical data has many advantages, implementation of similar methods with real-time data brings additional constraints and challenges. One such challenge is real-time data cleaning which would otherwise be executed off-line and with potential user intervention. Another is the lifespan of the data, where the data may become obsolete or invalid over time due to process, equipment, or tooling changes.

The selection of data acquisition sources should be based on both the availability of historical and real-time data, as well as the objectives that are to be achieved by using it, with an understanding of the constraints of each source. It may also be prudent to revisit data acquisition during solution formulation to evaluate the return on investment of refining data acquisition. While a fixed dataset may be used to formulate a Solution, a parallel analysis of the impact of adding additional data sources to the dataset based on successes illustrated in use case reports or general trends in usage may yield better results than committing to a fixed dataset. However, adding additional data sources changes the definition of the Problem and inherently affects the compatibility and interoperability between the Problem and Solution, so tradeoffs and downstream consequences should be taken into account when considering data acquisition changes. However, manufacturers often don't enjoy the luxury of being able to refine data acquisition methods and equipment, and must make the best of available data. This constraint can be driven by the inability to modify or upgrade a purchased machine tool, or the lack of resources to modify or upgrade custom equipment, which disproportionately affects smaller manufacturers.

According to the Department of Defense, small and medium manufacturers with less than 500 employees comprise nearly 99% of the approximately 291,000 manufacturers in the United States [22]. Due to their focus on lean production rather than emerging technologies for competitive advantage, and considering many manufacturing processes and related machinery at small manufacturers are designed and developed in-house based on closely held trade secrets and with limited engineering resources, this results in the use of the bare minimum required automation, sensors, digitalization, and future-proofing required to execute manufacturing processes. As a result, sensor selection is not tailored for ML or smart manufacturing and data acquisition refinement is likely required to achieve target objectives which benefit from ML methods. During data acquisition refinement or the digitalization of manual machinery, it is wise to take both the Problem-Solution compatibility with downstream consequences and examples of Solutions already applied to given data sources into account when incrementally upgrading. The sum cost of overhead and reconfiguration associated with each increment has the potential to balloon and exceed the cost of upgrades themselves, so consideration and attention should be given to the size of or minimization of the quantity of increments.

An area of interest for future research is ML applications in acoustic response (i.e., noise signature) and Acoustic Emission (AE), with acoustic response being within the range of human hearing and AE including frequency ranges extending in excess of audible sound. In the context of a cutting (machining) operation, the phenomena which generates AE is the dislocation in the primary shear zone and the sliding friction in the secondary shear zone [25]. Due to AE methods analyzing signals in a higher frequency range than machine vibrations and environmental noises, a distinct advantage is a relatively uncontaminated signal when combined with signal conditioning and processing, such as but not limited to a high pass filter [25]. While non-ML AE has been studied extensively in applications such as tool wear monitoring [24][25] and some ML AE applications have been examined such as in Fused Filament Fabrication (FFF) [149] and magnetic Barkhausen noise emission [142], the topic in a ML context is likely still ripe for discovery.

A constraint and consideration for AE applications is the volume of data that is generated due to sampling rates such as 500 kHz (circa 1993) [24], 10 MHz (circa 2019) [149], and 4 MHz (circa 2021) [142], as well as downstream requirements of ADC, signal conditioning and processing, data storage, and compute hardware necessary to facilitate AE and AE with ML. Due to sensor selection during machine design traditionally not considering AE methods and inherent sample rate limitations of typical control hardware which traditionally favor slow and robust operation, AE faces challenges for practical implementation on the factory floor, especially for legacy equipment. In addition, it is important to decouple and make the distinction between the phenomena which generates AE, the bandwidth of a sensor, the frequency range being analyzed, and methods of analyzing AE signals, especially during hardware selection. Analysis methods employed for AE are not inherently tied to specific phenomena, sensors, or frequency ranges, and selecting sensors with a large bandwidth and control or data acquisition hardware capable of capturing higher frequency ranges enables the use of AE analysis methods if phenomena which generate AE exist and can be adequately isolated from machine vibrations and environmental noise. Since experienced operators exhibit the capability of identifying process idiosyncrasies via audible indicators, which indicates that valuable information exists in lower frequency ranges than AE methods typically analyze, it stands to reason that ML may provide the capability of overcoming confounding in contaminated signals at lower frequencies and replicate experienced operator analysis. As a result, ML may present an option for side-stepping challenges related to isolation in lower frequency ranges which limit AE methods, thereby expanding analyzable frequency ranges and by proxy widening the field of acquirable insights.

*4.1.2. Decisions*

To be thorough, decisions made downstream from the ontology should be evaluated in tandem with data acquisition. Some common sense questions can be asked before engaging further steps to avoid misallocation of resources:

- In the possible permutations of outcomes or best case outcome of a decision made based on gained insights, do the outcomes justify the investment in resolving the Problem?

- If a Solution is formulated which resolves the Problem, are there barriers to deployment or implementation which need to be removed?

If any of these questions raise concern, focus should likely be placed on those areas prior to engaging further steps.

## 4.2. Step 2: Problem Formulation

After domain boundaries have been surveyed, Problem formulation can begin. As laid out in the Ontology section, the Problem domain consists of acquired data, desired insights, and a problem statement describing the transformation of data into insights and the relationship between them, without defining how that transformation takes place.

When formulating the problem statement, the level of abstraction at which the problem statement is defined should be commensurate with a detailed problem solving procedure. Since the quality of the formulated Solution is dependent upon the quality of the problem statement, defining a specific and precise problem statement is imperative. For example, a problem statement focused on technical outcomes such as detecting anomalous process parameters is significantly more valuable for Solution formulation than generic problem statements which represent business outcomes such as downtime reduction or profit optimization. While business success is the end goal, a generic problem statement does not address the nuts and bolts of how to get there.

Furthermore, separating nested and compound problems is preferable to isolate and modularize Solutions, unless a set of Problems requires a dependent or coupled solution. For instance, event detection may be a component of a process optimization Problem. By separating this into a Problem of event detection and a Problem of process optimization, separate components of the resulting solutions can be developed and understood independently, as well as reused independently. In addition, decoupling compound problems such as those represented by "Change Detection, Diagnostics, and Prognostics" problems detailed in Table 5 allow each Solution to be developed and understood independently if detection of a change will trigger the diagnostic or prognostic analysis in a sequence of tasks. However, if the detection of a change and the diagnostic analysis are dependent on each other, they may not be separable.

## 4.3. Step 3: Solution Formulation

Solution formulation is a complex topic, and is currently more of an art than a science. Due to the avant garde nature of Solutions, we cannot provide step by step guidance on the formulation process, and instead highlight areas which should be considered during formulation. However, one assumption underpins these considerations: prior to beginning Solution formulation, the data and insights defined in the Problem are fixed. Since the compatibility between Problem and Solution domains are defined by the data and insights, a change to the Problem definition would break the compatibility of solutions developed for that problem. Alternatively, Problems could be viewed as immutable, so any changes to the definition of a Problem would result in a new Problem which is incompatible with already constructed Solutions. Regardless of which perspective is employed, the Problem definition can be viewed as design constraints for Solution formulation.

In addition, making a distinction between Solution formulation and Problem resolution is necessary. It is possible to formulate Solutions which do not resolve the Problem, and compatibility between a Problem and a Solution does not infer that their union will produce desirable results. The determination of whether a Solution resolves a Problem is based on whether the Solution surpasses a specified performance threshold, which can be defined in the problem statement, should be defined before Solution formulation begins, and should be predicated on objective metrics.

### 4.3.1. Sequence Formulation

Sequence formulation is the process of formulating a sequence of tasks as illustrated in Figure 7 which define the structure of a Solution as illustrated in Figure 6. These tasks can consist of non-ML and ML Algorithms, and the order of operations are usually structured such that preprocessing tasks happen before tasks which generate insights. The order of formulation and dependencies may not be intuitive in a chicken and egg situation. For instance, specific preprocessing may be formulated to facilitate the use of a target Algorithm, or the use of a specific Algorithm may be facilitated by the use of target preprocessing tasks. However, preprocessing is one of the most important and time-consuming steps in every ML project., and it can have more impact on the prediction result than Algorithm selection. In the case of Algorithms which do not have the ability to appropriately handle missing values, preprocessing can provide a tradeoff which enables their use. Likewise, preprocessing can drive significant differences for Algorithms which have trouble handling non-gaussian features, non-normalized data, and high dimensionality. Conversely, a lack of preprocessing almost always negatively impacts results.

Based on the Problem and data complexity, a Solution consisting of one or several tasks should be constructed. Tasks are building blocks of the Solution and can be combined in different ways with respect to each other. Each task generates additional data or a contribution toward the global insights (see Figure 6 and Figure 7).

### 4.3.2. Preprocessing: Data Visualization

Before formulating any tasks in the Solution, we need to visualize the data and get a sense of what the available data looks like (e.g., distribution, missing values). This will help to design a suitable Solution which minimizes undesired bias and the loss of potential discriminatory information. In the case of sufficiently large real-world manufacturing data sets, visualization without dedicated tools is challenging. A large number of instances and the high dimensionality brings common tools used in manufacturing companies (e.g., MS Excel) quickly to their limits. The data visualization problem occurs in two primary instances: (i) limitations of the software to handle large data sets and (ii) high dimensionality exceeding the capacity of human comprehension. There are several tools and algorithms that can alleviate the problem of handling large data sets as well as addressing the transformation problem to make high dimensional data sets comprehensible. One of the latter is T-distributed Stochastic Neighbor Embedding (T-SNE). T-SNE is one of the few algorithms that are capable of visualizing high-dimensional data by projecting them into a low-dimensional (2D or 3D) space while capturing the underlying data properties. T-SNE is a non-linear data visualizer and the core idea behind it is the fact that similarity in high dimensions corresponds to a short distance in low dimensions. [107] adopted T-SNE for multivariate time-series and proposed m-TSNE to better understand MTS data. [148] proposed using two temporal matching measures, namely dynamic time warping (DTW) and angular metric for shape similarity (AMSS), to adopt T-SNE for time-series data visualization and enhance its ability.

### 4.3.3. Preprocessing: Feature Engineering

Instead of feeding raw data to the algorithm, feature engineering techniques can be used to enhance Solution quality. Feature Engineering is the process of using domain knowledge to extract, transform, or select relevant features (characteristics, properties, attributes) from raw data and to grasp the essence of the data in a form that is beneficial for downstream processing. It is the process of finding a set of features, or data characteristics that will most efficiently or meaningfully represent the data [140]. Feature Engineering can be divided into Feature Extraction, Feature Transformation, and Feature Selection which will be briefly discussed subsequently.

Feature extraction is a group of processes where a new set of non-redundant features are derived from existing features in the raw data. The goal of a feature extraction process is to reduce the data points while keeping the embedded information in the data and transforming the input parameter vector into a feature vector and/or reducing its dimensionality. In Time-series data, feature extraction tasks can be implemented within individual time-series (T) for the purpose of reducing the number of unnecessary points as well as removing autocorrelation within individual time-series. Based on Figure 1, T is the length of an individual time-series, M is the time-series input dimension and N is the number of observations. Piecewise Aggregate Approximation (PAA) [56] and Symbolic Fourier Approximation (SFA) [126] are examples of feature extraction algorithms with applications in time-series classification tasks.

Feature selection is the process of selecting a subset of features without changing or transforming them with the intent of removing redundant or irrelevant data. If the input data is a multivariate time-series it can be executed to identify and reduce the correlation between the dimensions (i.e., sensors) or change the data into a univariate time-series. It is worth noting that in manufacturing, the amount of needed data is highly related to the Problem, and implementing feature extraction without enough domain knowledge may result in inadvertently discarding useful information. For example in [36], Fulcher and Jones's feature-based linear classifier (FBL) algorithm generates a huge number of possible features from the input time-series and then it will filter them with a greedy feature selection mechanism to find the best features.

All types of feature engineering either increase or decrease the quantity of features, illustrated as T and M in Figure 1. The term Dimensionality Reduction is commonly used interchangeably with Feature Extraction when the intent is to focus only on extracted features which imply a decrease in the quantity of features. However, a more nuanced perspective reveals that the colloquial use of feature extraction implies feature selection, and is actually a two-step process. The first step in feature extraction is to extract a set of new features from a set of existing features, resulting in a master set of features which is a combination of the new and existing features. The quantity of new features is typically less than the quantity of existing features. The second step is to engage in feature selection, during which a subset of the master set is selected and used downstream, and selection is usually predicated on the assumption that the existing features contribute significantly less to the prediction outcome compared to the newly extracted features. This prediction outcome can be accuracy or other metrics which can be optimized such as computational expense. During the colloquial use of feature extraction, it is implied that the features selected in the second step are the new set of features extracted in the first step, and the existing features are discarded. For this reason, and since the total number of features has been reduced as a result of engaging both steps, the terminology of dimensionality reduction is used. However, when including the existing features, the total number of features has increased. Most importantly, a practitioner is not required to follow this convention and use the second step and may benefit from using the master set of features rather than only the newly extracted features.

Feature transformation is the process of the modification of raw data into features that are easier to understand by the ML Algorithm. Similar to feature extraction, this can be colloquially viewed as a two-step process where in the second step, the existing features are discarded in favor of the newly transformed features. There are different types of transformation techniques in the time-series analysis literature. Using Wavelet Transform algorithms such as discrete wavelet transform (DWT) in [110] and continuous wavelet transform (CWT) in [40], Fourier transform algorithms like fast Fourier Transform (FFT) in [94] and [63] and discrete Fourier transform (DFT) in [75], Hilbert-Huang transform (HHT) algorithm in [82], and Gramian Angular Fields (GAF) algorithm in [89] are a few examples of feature transformation techniques that are designed to transform the data from the time domain into other domains (e.g. frequency, image, etc).

An additional and conceptually different feature engineering process, which is also commonly referred to as feature selection, focuses on the reduction of the time-series length T (or sub-features) illustrated in Figure 1 within a single feature with the intent of removing redundant or irrelevant data. This can be accomplished through a variety of methods such as downsampling, moving average, and smoothing. When excessive sampling rates are used or redundant data is encountered and computational power or storage costs are limiting factors, this process can be used to alleviate these concerns. However, whether sampling rates are excessive is relative to the manufacturing process, the features being analyzed, and the objective of the analysis. For instance, a sampling rate near the tooth pass frequency of a vertical mill may be adequate for detecting tooth impact frequency and magnitude, but would not be adequate for capturing transients and the acoustic signature of the system which would require significantly higher sampling rates and the analysis of which might produce completely different results. The nuances of tradeoffs between resource consumption and analysis capability, as well as assumptions about what data and methods of analysis are of value, should be thoroughly examined and understood on a case by case basis prior to applying this process. The core advantage of ML over traditional methods is its ability to find non-intuitive correlations between large amounts of input and output data, and this process always carries the risk of discarding useful data which would otherwise enable or improve the ability to capitalize on this core advantage.

### 4.3.4. Preprocessing: Input Dimensionality

Input data dimension M (i.e., univariate time-series vs multivariate time-series) is an important parameter that should be investigated in this step. All feature engineering tasks that were discussed in the feature extraction section can also be implemented on the input data dimension. The main additional complexity for multivariate time-series is the discriminatory interactions between dimensions (i.e., sensors), and not just in the autocorrelation within an individual time-series [121]. Despite the fact that many industrial problems are inherently multivariate (e.g., system monitoring), due to the complexity of dealing with multivariate time-series, much less consideration has been given to multivariate problems in the literature compared to univariate problems. Furthermore, many ML algorithms are designed only for univariate time-series and are incapable of dealing with multivariate time-series. There are several techniques to go from multivariate time-series back to univariate time-series. One way is simply by ensembling over dimensions and assuming independence between dimensions [121]. We can alleviate this strong assumption with some techniques such as Principal Component Analysis (PCA). PCA aims at mathematically choosing orthogonal features as principal components as a preprocessing step or making use of class labels as the teacher and controlling mechanism and assume that we would lose information and consequently accuracy if made the wrong assumption. PCA works by constructing new artificial variables (principal components) which are linear combinations of original variables. The newly constructed variables are orthogonal to each other and they can represent the data via linearly independent vectors.

When considering input dimensionality during the Algorithm selection process, there are two options for implementing a Solution using multivariate data: (i) using an Algorithm in a task which accepts data from a multivariate data source, and (ii) using multiple univariate Algorithms as separate tasks in parallel and combine their output data or insights in a downstream task.

### 4.3.5. Algorithm Selection: Approach

During the Algorithm selection process, multiple Approaches may be compatible with output dimensionality and label constraints, and just because labels are available does not inherently mean that a practitioner must or even should use them. Approaches which have a less stringent label requirement may be substituted to drive strategic outcomes. For instance, SL, SSL, or UL can be applied to a fully labeled data set, only SSL or UL can be applied to a partially labeled data set, and only UL can be applied to an unlabeled data set. Additionally, multiple Algorithms can be employed using different Approaches in parallel to characterize performance or compare derived insights. For instance, the performance of SL, SSL, and UL can be compared in parallel on a fully labeled data set, the performance of SSL and UL can be compared in parallel on a partially labeled data set, and only UL can be used on an unlabeled data set. While multiple Approaches may be applicable in a given situation, it is important to understand the underlying assumptions and implicit biases of each Approach.

When fully labeled data is available and you are willing to commit to the assumption that the labels are accurate, SL has several advantages over alternatives and is preferred in most cases. First, it can be significantly more accurate than alternative Approaches. Additionally, it enables an improvement over UL for visualization methods through dimensionality reduction. A disadvantage of SL is the increased cost if a subject matter expert has to manually label the data.

While labeled data may be readily available and provide a clear avenue toward SL, the accuracy and trustworthiness of labels should be considered before committing to a SL approach. In manufacturing, data can come from a wide variety of sources and with varying degrees of integrity. It is important to understand the chain of custody of data, how it is acquired, and what intermediate processing is taking place, including whether labels are synthesized from the data itself during the data acquisition process. This is especially true with respect to data derived from sensors, where the label may be synthesized from sensor data at the PLC or control level. Without this understanding, practitioners run the risk of redundant training, where labels are synthesized from the sensor data, then the labels and sensor data are used to train the model, effectively using the same data on both sides of the model during training. Since there are no explicit indicators of this occurring and a practitioner may not have access to or an understanding of the PLC or control system code to verify, this could easily go undetected or cause issues which are challenging to diagnose the source of. Unchecked and undiagnosed poor or overinflated accuracy resulting from this is likely to adversely impact decision making processes and downstream system performance.

Regardless of whether the validity of labels in a fully labeled dataset can be verified, an UL learning approach can provide an avenue for the exploration of insights not discoverable through SL and therefore offer a strategic alternative. In addition, insights may be gained through exploratory UL not attainable through SL, which may produce overall better outcomes. Alternatively, labels can be used to validate UL clusters and detect outliers.

A challenge when dealing with partially labeled data is choosing an appropriate solution, of which there are typically two methods which are considered Semi-Supervised Learning (SSL). The first method is to use a UL algorithm to generate labels, then use those generated labels to train a SL algorithm using the same input data, in a sequence of tasks. This method, entitled Semi-Supervised Classification, leverages UL to enable the use of SL methods that otherwise could not be employed [165]. With a number of implicit assumptions and tradeoffs, this is executed by utilizing labeled data to construct a decision boundary for predicting labels both on future test data and unlabeled data, which can be coupled with UL clusters to synthesize labels for unlabeled data before executing SL [165]. The second method is to use a composite algorithm combining SL and UL, where the SL algorithm guides or constrains the learning of the UL algorithm. This method, entitled Constrained Clustering, uses SL to enhance UL through the knowledge that some data points belong to a specific cluster, and is a composite algorithm [165][27]. When label availability is not the driving constraint, SSL carries the distinct advantage of reducing a subject matter expert's required labor of manually labeling data in order to take advantage of SL Approaches. However, successful implementation is likely to correlate to whether all clusters, including edge cases, are adequately labeled, and assumes that no unknown states exist. By extension, SSL could enable the use of SL methods on extremely large datasets where the cost of manually labeling the entire dataset is prohibitively expensive. Within the reviewed papers, SSL applications combined a small amount of labeled data with a large amount of unlabeled data [27][165].

### 4.3.6. Algorithm Selection: Algorithms vs Models

$$\text{Algorithm} + \frac{\text{Training}}{\text{Parameters}} + \frac{\text{Training}}{\text{Data}} = \text{Model}$$

**Figure 28:** Model Equation

During the Algorithm selection process, it is important to distinguish between an ML algorithm and an ML model. An algorithm is a process or sequence of operations which is completely independent of the data it is operating upon and may have parameters which can be optimized to produce a desired result. A model is the result of a training process where an algorithm operates on a dataset with a specified set of parameters. During this process, the algorithm, training parameters, and training data are combined to form the model, so the resulting model is implicitly biased by these three addends, as illustrated arithmetically in Figure 28.

$$\text{Model} + \frac{\text{Test}}{\text{Data}} = \text{Performance}$$

**Figure 29:** Performance Equation

After a model has been produced by the training process and during application in industry or during the performance evaluation of a model, the performance of the model is quantified by metrics such as accuracy. The performance of the model is a combination of the test data and the model itself, as illustrated arithmetically in Figure 29. When Figure 28 is substituted into Figure 29, Figure 30 gives a clear perspective on the four addends contributing to performance.

$$\text{Algorithm} + \frac{\text{Training}}{\text{Parameters}} + \frac{\text{Training}}{\text{Data}} + \frac{\text{Test}}{\text{Data}} = \text{Performance}$$

**Figure 30:** Performance Equation Expanded

Russel et al defined Transfer Learning (TL) as enabling models to be trained on a source domain task and be applied to a separate target domain task [135]. In this definition we interpret "domain" to be equatable to a Problem under the proposed ontology or a problem outside of manufacturing, and successful TL would equate to exceeding the threshold for determining that a Solution resolves a Problem other than Problem it was originally intended to resolve. However, exploring strategically incorrect interpretations of this definition reveals valuable nuances related to the Algorithm selection process.

First, there is the notion of transferability into the future, or Process Time Transferability (PTT), where a fixed model (fixed algorithm, training parameters, and training data) which was trained on a machine is executed on the same machine, but at two different points in time. In this case, the test data is the only visibly changing variable in Figure 30. However, the relative relationship between training data and test data may become a factor when considering training data lifespan and downstream model lifespan, being the amount of time during which it remains valid and accurately reflects the manufacturing process. There are a variety of potential causes for shortened training data lifespan, such as machine wear and degradation over time. In the case of the replacement of wear components in a machine (e.g. replacing or sharpening tooling), a small training dataset which does not include multiple replacement cycles would theoretically be at higher risk. A more complex case is in changing material lots, where changes are cyclical but raw material properties are unique and constantly evolving. The level of sensitivity to these changes should be a driver for depth of study, which could include drilling down to the chemical composition and processing parameters of the raw material and how that affects the manufacturing process at hand. When selecting an Algorithm, the limitations and lifespan of training data, as well as how well a given Algorithm tolerates or is sensitive to them, are important considerations.

Second, there is the notion of transferability across manufacturing process instances, or Process Instance Transferability (PIT), where a fixed model (fixed algorithm, training parameters, and training data) which was trained on machine A is executed on machine B, and where both machines were built to the same specifications but differ due to variability in the manufacturing processes which produced the parts which comprise the machines. In this case, the test data is the only changing variable in Figure 30. However, the masked limitation is that there may be a difference of significance between Training Data derived from each machine, which when compounded with the difference in Model Data, may extrapolate into worse performance. An alternative method is to use the same algorithm and training parameters, but use Training Data from the instance of the machine which it will be tested on. This individualizes performance and sidesteps instance-driven performance differences, but adds data management constraints which favor MLOps or DevOps methodologies. Yet another alternative would be to choose an Algorithm which is less sensitive to or can tolerate the differences between instances.

Finally, there is a fallacy of transferability of Algorithm performance. In this case, a practitioner might attempt to apply an Algorithm which resolved the same Problem for another practitioner using a different training dataset. If we want to grant that the same test data is being used for the sake of comparison, the practitioner might find that the performance is better or worse. If worse and the formulated Solution does not resolve the Problem, the practitioner might attempt to use a different Algorithm to resolve the Problem. However, the jump to this decision indicates that the variable of training parameters was overlooked. There may be a need to compensate for differences in training data by optimizing training parameters to produce a model which achieves similar or better performance.

Since the biases of the data are inherited by the model during the training process and the performance of the model is dependent on a combination of both the Algorithm and data being operated on, the performance of a model is not inherently transferable to another dataset, nor is it representative of the Algorithm's ability to produce a model which performs better through optimization. The training parameter optimization process has the potential to become an exorbitant resource drain. Therefore, it is important to define the scope of exploration and understand implicit assumptions before selecting a family of Solutions to examine and apply to a Problem based on the results of other practitioners.

### 4.4. Step 4: Performance Evaluation

The final step is to evaluate whether and how well the Solution resolved the Problem. The topic of evaluation metrics did not receive significant attention or discussion in the reviewed literature. Since evaluation metric values and analysis were sparse in the reviewed literature, we deemed the topic outside the scope of this review. However, given its importance, we included this step as a placeholder for future research and exploration.

## 5. Conclusions

This paper provides an overview of time-series pattern recognition in manufacturing. It encompasses related literature reviews, our motivation for writing this review, the scope of the review, and details the method by which it was executed. It also proposes an ontology to simplify, contextualize, and provide a consistent and cohesive structure for the results. In addition, it contains a proposed procedure for selecting Algorithms while highlighting the need for an objective method for classifying Algorithms, and it intentionally detaches subjective terminology from underlying concepts to illuminate distinctions and nuances. Furthermore, this provides results in the form of detailed reference material and discussion targeting insights for practical application with reference to additional literature through which more information can be ascertained. Finally, this paper provides high-level, novel guidelines for and considerations when leveraging the results presented.

However, there are a number of topics which were outside the scope of this paper or not explored in depth which we feel warrant further investigation. While the concept of compatibility and interoperability has been discussed, a method of enabling or enforcing them, such as standards, were not addressed. The topic of RL was also touched on but not explored in depth due to the lack of application in manufacturing to date, and the underlying reasons why and projected growth in the future are of interest. Similarly, Techniques which have been applied in the computer science field but not in manufacturing are ripe for exploration, along with the reasons why they have not been explored to date. Feature Engineering techniques and algorithms for time-series data are another topic that deserves more attention from researchers moving forward. Although we briefly scratched the surface, a comprehensive review of the topic is deemed out of scope for this review and can be considered an interesting topic for future research.

Metrics were also discussed along with the benefits of and justifications for them, but there was no analysis of which metrics are of merit, how the value of specific metrics correlate to a Solution's ability to resolve a Problem, how metrics can be properly leveraged for benchmarking, or whether metrics which provide merit for benchmarking correlate to the metrics which provide merit when determining a Solution's ability to resolve a Problem. Besides accuracy, metrics such as cost warrant special attention with respect to energy consumption and resource utilization which can drive return on investment analysis to support business decisions. The current lack of experience with successfully monetized AI applications in manufacturing beyond individual operations is a limiting factor [141] and both the justifications for and visibility of returns on investment are prerequisites to Solution implementation. Similarly, developing an objective metric to analyze the similarity of Algorithms and Solutions based on structure and functionality would improve the proposed Algorithm selection procedure. In tandem, an analysis of Algorithm and Solution genealogy might yield similar insights which aid in Technique classification. Most importantly, we use the term "performance" rather than "accuracy" to denote that accuracy might not be the most important indicator of a desirable outcome. In manufacturing specifically, metrics such as energy consumption or cost may be the parameters which are most important and being optimized for, as long as the accuracy of the Solution meets the threshold of *good enough* to resolve the Problem.

### 5.1. A Vision for the Future

While these conclusions provide a few topics which warrant further investigation, a core missing component is an overarching vision for the future. Time-series analytics in manufacturing is currently a skillfully executed and handcrafted art. The transformation from an art to a science consisting of rigorous methods for Problem and Solution formulation by simplifying and bottling the knowledge of experts is critical to driving the democratization of ML in manufacturing. A component of this transformation is the formulation of the basic building blocks through which compatibility and composability can be enabled and Solutions modularly assembled. Once compatibility has been established, tools can be developed to automate formulation steps which facilitate an exponential increase in the rate of research using less manpower until computational resource saturation is reached. While increasing the volume of research through automation does not guarantee quality results or efficient use of resources, it does provide a fast route toward results which may prove *good enough* in a short calendar time frame.

Furthermore, if the objective structure of Algorithms and Solutions are studied and mapped, this could be extended to automate the exploration of new Algorithm and Solution structures. While the ability to democratize consumption of existing domain knowledge, Algorithms, and Solutions can be enabled by the transition from an art to a science, the democratization of the creation of new Algorithms and Solutions is dependent on compatibility, automation, and analyzability based on objective metrics which serve as a useful imposter for deep domain knowledge. In manufacturing, where Solutions need only be *good enough* to resolve a Problem rather than finding the best possible Solution, the strategic combination of democratization and The Infinite Monkey Theorem may be the most expedient way to advance time-series analytics in manufacturing.

### 5.2. Recommendations

To facilitate the pursuit of this vision for the future, the reluctance to share data and detailed use case reports must be recognized as major roadblocks to the advancement of time-series analytics in manufacturing. The sharing of data and use case reports enables the refinement of the proposed ontology and standards which underpin common repositories for shared data in standardized formats, standard metrics and datasets for benchmarking, and more expansive domain knowledge and reference material to drive decision making such as the materials found in the Appendix.

The formulation of guidelines and analysis of use cases is currently limited by the availability, structure, and specificity of use cases reports. In two cases, Algorithms could not be included in our analysis because the Algorithms were not specified in enough detail to effectively extract data, which may be in part due to using commercial software for time-series analytics [109][123]. To facilitate the future extraction of thorough and detailed information, Figure 31 details a structure for organizing and sharing data. By using it as a guideline for the minimum information to share in a use case report or publication, further analysis similar to the results presented in this paper are enabled. In addition to expanding the analysis presented in this paper, deeper analysis is enabled through the inclusion of directed graphs to describe the structure of sequences of tasks as and the structure of composite Algorithms, which might look similar to the use case presented in the Use Case Example section. However, more detailed and specific information would be prefered over this bare minimum, such as what data and insights are flowing between tasks in Figure 13, to facilitate deeper analysis as well as understand where our ontology and proposed data structure break down and need refinement. In addition, the inclusion of metrics enables performance analysis and comparison of solutions, while specifying a publicly available data set enables objective apples to apples performance benchmarking.

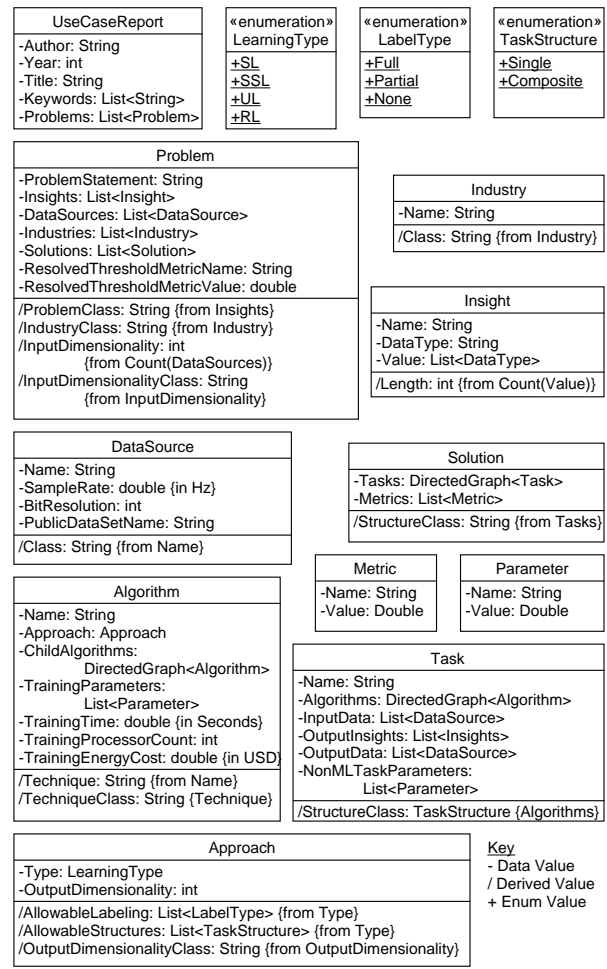

**Figure 31:** Proposed Data Structure

### 5.3. Limitations

This paper was written under certain assumptions, timelines, and resource limitations and is based on our understanding of the subject matter with the intent of clearly articulating and following the process and methodology we laid out. While papers in this field are typically focused on objective and quantitative analyses and we recommend an effort be made to move this topic further in that direction, much of the data and analyses in this paper is subjective and heavily biased by our perceptions, including the lens of problem-solving and practical applicability through which it was written. Since the complete removal of subjectivity and biases is not only impossible but arguably not desirable, our intent is to be transparent about the process and methodology that was followed such that our audience understands our biases, intent, understanding, and their influence on the content of this paper.

### 5.4. Acknowledgment

This material is based upon work supported by the National Science Foundation under Grant No. 2119654. Any opinions, findings, and conclusions or recommendations expressed in this material are those of the author(s) and do not necessarily reflect the views of the National Science Foundation.

The authors express their appreciation for the encouragement and valuable discussions among the greater Smart Manufacturing community that improved the quality of this paper. In particular, the authors thank Dr. Jim Davis and Dr. John Roth for their inspiration, and Dr. Eamonn Keogh and the Journal of Manufacturing Systems reviewers for their feedback and resulting improvements.

# A. Appendix

| Technique | Class | Algorithm | Univariate | Multivariate |
|---|---|---|---|---|
| Instance-Based (16) | Conventional | K-Nearest Neighbor (KNN) (5) | [87] [139] | [29] [16] [105] |
| | | Support Vector Machines (SVM) (5) | [64] [130] [139] | [2] [105] |
| | | One-Class SVM (3) | [70] | [105] [63] |
| | | Dynamic Time Warping (DTW) (1) | | [52] |
| | | Support Vector Data Description (SVDD) (1) | | [62] |
| | | Learning Vector Quantization (LVQ) (1) | [130] | |
| Recurrent Neural Networks (RNN) (13) | ANN/DL | LSTM Network (7) | [75] [54] [132] [33] | [51] [19] [73] |
| | | CNN-LSTM Network (2) | [72] [72] | |
| | | BDLSTM (Bi-LSTM) network (2) | [160] [38] | |
| | | Gated Recurrent Unit (GRU) (1) | | [19] |
| | | RNN (1) | | [73] |
| Convolutional Neural Networks (CNN) (10) | ANN/DL | Convolutional Neural Network (CNN) (6) | [97] [124] [75] [38] | [58] [47] |
| | | 1D-CNN (2) | [122] [98] | |
| | | Fully Convolutional Networks (FCN) (1) | | [147] |
| | | Temporal Convolutional Networks (TCN) (1) | | [51] |
| Decision Tree-Based (8) | Conventional | Random Forest (RF) (6) | [97] [64] [32] | [82] [4] [16] |
| | | eXtreme Gradient Boosting (XGBoost) (1) | [64] | |
| | | Gradient Boosting Classifier (GBM) (1) | [64] | |
| FF Neural Networks (8) | ANN/DL | Multi Layer Perceptron's (MLP) (4) | [132] [11] | [50] [2] |
| | | Fully Connected Neural Networks (FCNN) (2) | [97] | [71] |
| | | Back Propagation Network (BPN) (2) | [95] | [76] |
| Statistical (6) | Conventional | Linear Discriminant Analysis (LDA) (2) | [139] | [16] |
| | | Quadratic Discriminant Analysis (QDA) (2) | [139] | [16] |
| | | Naive Bayes Classifier (2) | | [29] [16] |
| Regression (4) | Conventional | Logistic Regression (LR) (4) | [97] [112] [64] | [2] |
| Encoder/Decoder Based (3) | ANN/DL | Variational AE (VAE) (1) | | [61] |
| | | Deep Stacked Autoencoder (DSAE) (1) | [95] | |
| | | Attention Mechanism (1) | [160] | |

**Table 13:** BSL Algorithms

| Technique | Class | Algorithm | Univariate | Multivariate |
|---|---|---|---|---|
| FF Neural Networks (25) | ANN/DL | Multi Layer Perceptron's (MLP) (14) | [26] [137] [113] [114] [115] [30] [162] [88] [20] | [145] [155] [133] [91] [28] |
| | | Back Propagation Network (BPN) (4) | [144] [60] [156] [18] | |
| | | Radial Basis Function Neural Networks (RBFNNs) (2) | [119] [26] | |
| | | Fully Connected Neural Networks (FCNN) (2) | | [10] [28] |
| | | Feedforward Fully Connected Spiking Neural Network (1) | [5] | |
| | | ESSM (1) | [161] | |
| | | Probabilistic Neural Networks (PNNs) (1) | [26] | |
| Decision Tree-Based (21) | Conventional | Random Forest (RF) (11) | [102] [78] [20] [93] [103] [101] | [41] [59] [120] [28] [67] |
| | | eXtreme Gradient Boosting (XGBoost) (3) | [103] | [120] [10] |
| | | Gradient Boosting Classifier (GBM) (3) | | [41] [59] [120] |
| | | Decision Tree (Dtree) (1) | [103] | |
| | | Bagging Classifier (BAG) (1) | | [41] |
| | | RUS Boosted Decision Trees (1) | | [63] |
| | | Extreme Random Forest (ERF) (1) | [20] | |
| Convolutional Neural Networks (CNN) (19) | ANN/DL | Convolutional Neural Network (CNN) (11) | [142] [116] [110] [37] [100] [101] | [89] [43] [41] [28] [1] |
| | | 1D-CNN (3) | [154] [159] | [152] |
| | | Residual Neural Network (ResNet) (2) | | [145] [155] |
| | | HHO-ConvNet (1) | [39] | |
| | | FDC-CNN Model (1) | | [74] |
| | | Explainable DCNN (1) | [40] | |
| Recurrent Neural Networks (RNN) (14) | ANN/DL | LSTM Network (5) | [102] [80] | [155] [94] [28] |
| | | CNN-LSTM Network (2) | | [94] [94] |
| | | Gated Recurrent Unit (GRU) (2) | | [81] [10] |
| | | BDLSTM (Bi-LSTM) network (2) | [158] | [8] |
| | | NARX Neural Network (1) | [104] | |
| | | AtLSTMs (1) | | [155] |
| | | Stacked LSTM (1) | | [94] |
| Instance-Based (12) | Conventional | Support Vector Machines (SVM) (6) | [116] [137] [20] | [120] [41] [67] |
| | | K-Nearest Neighbor (KNN) (4) | [116] [101] | [41] [28] |
| | | Self-Organizing Map (SOM) (1) | [119] | |
| | | Learning Vector Quantization (LVQ) (1) | [156] | |
| Statistical (2) | Conventional | Naive Bayes Classifier (2) | [101] | [41] |
| Encoder/Decoder Based (1) | ANN/DL | Stacked Denoising Autoencoder (SDAE) (1) | | [157] |

Table 14: MSL Algorithms

| Technique | Class | Algorithm | Univariate | Multivariate |
|---|---|---|---|---|
| Semi Supervised Detection of Outliers (SSDO) (1) | Conventional | COP K-Means Constrained Clustering Algorithm (1) | [143] | |

Table 15: BSSL Algorithms

| Technique | Class | Algorithm | Univariate | Multivariate |
|---|---|---|---|---|
| Encoder/Decoder Based (8) | ANN/DL | Encoder Decoder (E/D) (2) | [49] | [129] |
| | | RNN E/D (2) | [125] | [129] |
| | | Adversarial Autoencoder (AAE) (1) | | [14] |
| | | Transformer Encoder (1) | [96] | |
| | | ECHAD (Embedding-based CHAnge Detection) (1) | | [17] |
| | | Variational Bi-LSTM AE (1) | | [111] |
| Constraint-Based (4) | Conventional | Isolation Forest (iForest) (4) | [70] [23] | [68] [129] |
| Density-Based (4) | Conventional | Local Outlier Factor (LOF) (3) | | [129] [62] [68] |
| | | Kernel Spectral Clustering (KSC) (1) | | [69] |
| Center-Based (4) | Conventional | K-Means Clustering (3) | [23] | [29] [62] |
| | | Histogram-Based Outlier Score (HBOS) (1) | | [129] |
| Distribution-Based (4) | Conventional | Gaussian Mixture Models (GMM) (3) | [49] | [53] [62] |
| | | Parzen Window (1) | | [62] |
| Convolutional Neural Networks (CNN) (4) | ANN/DL | 1D-CNN (2) | [125] | [90] |
| | | Convolutional Neural Network (CNN) (2) | [49] [15] | |
| Dimension Reduction (3) | Conventional | Principal Component Analysis (PCA) (2) | [20] | [62] |
| | | Kernel PCA (KPCA) (1) | | [62] |
| FF Neural Networks (3) | ANN/DL | Fully Connected Neural Networks (FCNN) (2) | [49] | [90] |
| | | Bayesian Interpretation of Neural Networks (BINN) (1) | [15] | |
| Generative Adversarial Models (3) | ANN/DL | Generative Adversarial Network (GAN) (2) | | [129] [14] |
| | | Mode Seeking Generative Adversial Network (MSGAN) (1) | | [85] |
| Recurrent Neural Networks (RNN) (3) | ANN/DL | RNN (1) | [125] | |
| | | Stacked LSTM (1) | | [90] |
| | | CNN-LSTM Network (1) | [83] | |
| Connectivity-Based (2) | Conventional | Hierarchical TICC (1) | | [8] |
| | | Hierarchical Clustering (1) | | [50] |

Table 16: BUL Algorithms

| Technique | Class | Algorithm | Univariate | Multivariate |
|---|---|---|---|---|
| Center-Based (12) | Conventional | K-Means Clustering (9) | [128] [55] [46] [127] [26] | [120] [59] [13] [53] |
| | | K-MICA clustering (1) | [26] | |
| | | Kernel K-means (1) | | [79] |
| | | K-Medoids Clustering (1) | [46] | |
| Connectivity-Based (6) | Conventional | Agglomerative Hierarchical Clustering (AGNES) (2) | [118] [46] | |
| | | Hierarchical TICC (2) | | [145] [53] |
| | | Weighted Pair Group Method with Arithmetic Mean (WPGMA) (1) | [46] | |
| | | Spectral Clustering (1) | [46] | |
| Density-Based (3) | Conventional | MDL clustering (1) | [46] | |
| | | OPTICS (1) | [46] | |
| | | DBSCAN (1) | [46] | |
| Dimension Reduction (1) | Conventional | Self Organizing Map (SOM) (1) | [60] | |
| Distribution-Based (1) | Conventional | Gaussian Mixture Models (GMM) (1) | [46] | |
| Generative Adversarial Models (1) | ANN/DL | Generative Adversarial Network (GAN) (1) | | [152] |

Table 17: MUL Algorithms

| | | | Problem | | | | | | | | | | Input | | | |
|---|---|---|---|---|---|---|---|---|---|---|---|---|---|---|---|---|
| | | | Class | | | | Attribute | | | | | | Class | | | |
| | | | Event & State Analysis | Quality Control | Optimization | Total | Diagnostics & Prognostics | Change Detection | Quality Control | Change Detection, Diagnostics, & Prognostics | Process Optimization | Factory/Network Optimization | Total | Univariate | Multivariate | Total | |
| Class | General | | 31 | 19 | 14 | 64 | 14 | 15 | 19 | 2 | 8 | 6 | 64 | 40 | 24 | 64 | |
| | Industry-Specific | | 49 | | 4 | 53 | 26 | 15 | | 8 | 1 | 3 | 53 | 26 | 27 | 53 | |
| | Total | | 80 | 19 | 18 | 117 | 40 | 30 | 19 | 10 | 9 | 9 | 117 | 66 | 51 | 117 | |
| Attribute | General | General Manufacturing | 31 | 19 | 14 | 64 | 14 | 15 | 19 | 2 | 8 | 6 | 64 | 40 | 24 | 64 | |
| | Industry-Specific | Automotive | 9 | | | 9 | 5 | 3 | | 1 | | | 9 | 7 | 2 | 9 | |
| | Industry-Specific | Energy | 4 | | 3 | 7 | 1 | 3 | | | | 3 | 7 | 4 | 3 | 7 | |
| | Industry-Specific | Oil and Gas | 5 | | 1 | 6 | 1 | 2 | | 2 | 1 | | 6 | 3 | 3 | 6 | |
| | Industry-Specific | Semiconductor | 6 | | | 6 | 2 | 1 | | 3 | | | 6 | 1 | 5 | 6 | |
| | Industry-Specific | Healthcare | 6 | | | 6 | 6 | | | | | | 6 | 3 | 3 | 6 | |
| | Industry-Specific | Agriculture and Food | 5 | | | 5 | 4 | | | 1 | | | 5 | 1 | 4 | 5 | |
| | Industry-Specific | Aerospace | 5 | | | 5 | 3 | 1 | | 1 | | | 5 | 2 | 3 | 5 | |
| | Industry-Specific | Chemical | 4 | | | 4 | 2 | 2 | | | | | 4 | 2 | 2 | 4 | |
| | Industry-Specific | Construction & Utilities | 3 | | | 3 | 1 | 2 | | | | | 3 | 1 | 2 | 3 | |
| | Industry-Specific | Additive Manufacturing | 2 | | | 2 | 1 | 1 | | | | | 2 | 2 | | 2 | |
| | Total | | 80 | 19 | 18 | 117 | 40 | 30 | 19 | 10 | 9 | 9 | 117 | 66 | 51 | 117 | |

Figure 32: Industry-Problem Domain Matrix

| | | | Input | | |
|---|---|---|---|---|---|
| | | | Class | | |
| | Class | Attribute | Univariate | Multivariate | Total |
| Class | Event & State Analysis | | 40 | 40 | 80 |
| | Quality Control | | 16 | 3 | 19 |
| | Optimization | | 10 | 8 | 18 |
| | Total | | 66 | 51 | 117 |
| Attribute | Event & State Analysis | Diagnostics & Prognostics | 23 | 17 | 40 |
| | Event & State Analysis | Change Detection | 16 | 14 | 30 |
| | Quality Control | Quality Control | 16 | 3 | 19 |
| | Event & State Analysis | Change Detection, Diagnostics, & Prognostics | 1 | 9 | 10 |
| | Optimization | Process Optimization | 3 | 6 | 9 |
| | Optimization | Factory/Network Optimization | 7 | 2 | 9 |
| | Total | | 66 | 51 | 117 |

Figure 33: Problem-Problem Domain Matrix

**Figure 34:** Data Source-Problem Domain Matrix

| Jargon Attribute | Problem Class | | | Problem Attribute | | | | | | Industry Class | | | Industry Attribute | | | | | | | | | | | Input Class | | |
|---|---|---|---|---|---|---|---|---|---|---|---|---|---|---|---|---|---|---|---|---|---|---|---|---|---|---|
| | Event & State Analysis | Quality Control | Optimization | Event & State Analysis: Diagnostics & Prognostics | Event & State Analysis: Change Detection | Quality Control: Quality Control | Optimization: Change Detection, Diagnostics, & Prognostics | Optimization: Process Optimization | Optimization: Factory/Network Optimization | Total | General | Industry-Specific | Total | General Manufacturing | Automotive | Energy | Oil and Gas | Semiconductor | Healthcare | Agriculture and Food | Aerospace | Chemical | Construction & Utilities | Additive Manufacturing | Univariate | Multivariate | Total |
| Control Chart Pattern Recognition | 1 | 17 | 1 | 19 | 1 | | 17 | | 1 | | 19 | 18 | 1 | 19 | 18 | | | | | | | | 1 | | | 19 | 17 | 2 | 19 |
| Anomaly Detection | 16 | | 1 | 17 | 1 | 15 | | | | 1 | 17 | 8 | 9 | 17 | 8 | 3 | 1 | | | | 1 | 2 | 2 | | | 17 | 7 | 10 | 17 |
| Fault Detection | 12 | | 1 | 13 | 2 | 6 | | 4 | 1 | | 13 | 7 | 6 | 13 | 7 | 1 | 1 | 1 | 1 | | 1 | | | | 1 | 13 | 6 | 7 | 13 |
| Pattern Recognition | 9 | 1 | 3 | 13 | 8 | 1 | 1 | | 1 | 2 | 13 | 7 | 6 | 13 | 7 | 1 | 1 | 1 | | 2 | 1 | | | | | 13 | 8 | 5 | 13 |
| Time Series Classification | 5 | 5 | | 10 | 5 | | | 3 | 2 | | 10 | 5 | 5 | 10 | 5 | | | 1 | | 2 | | 1 | | | 1 | 10 | 7 | 3 | 10 |
| Fault Diagnosis | 6 | | | 6 | 6 | | | | | | 6 | 4 | 2 | 6 | 4 | 1 | | | 1 | | | | | | | 6 | 5 | 1 | 6 |
| Fault Detection and Diagnosis | 3 | | 1 | 4 | 1 | | 2 | 1 | | | 4 | 1 | 3 | 4 | 1 | | | | 2 | | | 1 | | | | 4 | 1 | 3 | 4 |
| Time Series Clustering | 1 | | 2 | 3 | 1 | | | | | 2 | 3 | | 3 | 3 | | | 2 | | | | 1 | | | | | 3 | 2 | 1 | 3 |
| Binary Classification | 1 | | 1 | 2 | 1 | | | 1 | | | 2 | 1 | 1 | 2 | 1 | 1 | | | | | | | | | | 2 | 1 | 1 | 2 |
| Fault Detection and Classification (FDC) | 2 | | | 2 | 2 | | | | | | 2 | 2 | | 2 | | | 1 | | | 1 | | | | | | 2 | 1 | 1 | 2 |
| Damage Detection | 1 | | 1 | 2 | | | | | 1 | | 2 | 1 | 1 | 2 | 1 | | | 1 | | | | | | | | 2 | 1 | 1 | 2 |
| Change Detection | 1 | 1 | | 2 | | | 1 | 1 | | | 2 | 1 | 1 | 2 | 1 | | 1 | | | | | | | | | 2 | 1 | 1 | 2 |
| Machine State Detection | 2 | | | 2 | 2 | | | | | | 2 | 2 | | 2 | 2 | | | | | | | | | | | 2 | 1 | 1 | 2 |
| Human Activity Recognition (HAR) | 1 | | | 1 | 1 | | | | | | 1 | 1 | 1 | 1 | | | | | | 1 | | | | | | 1 | | 1 | 1 |
| Earthquake Detection | 1 | | | 1 | 1 | | | | | | 1 | 1 | 1 | 1 | | | | | | | | 1 | | | | 1 | | 1 | 1 |
| Annode Effect (AE) Detection | 1 | | | 1 | 1 | | | | | | 1 | 1 | 1 | 1 | | | | | | | | | 1 | | | 1 | 1 | | 1 |
| Strip Breakage Detection | | 1 | | 1 | | | 1 | | | | 1 | 1 | 1 | 1 | | | | | | | | | | | | 1 | | 1 | 1 |
| Tool Wear Detection | 1 | | | 1 | 1 | | | | | | 1 | 1 | 1 | 1 | | | | | | | | | | | | 1 | 1 | | 1 |
| Unbalance Detection | 1 | | | 1 | | | | 1 | | | 1 | 1 | 1 | 1 | | | | | | | | | | | | 1 | 1 | | 1 |
| Disease Detection | 1 | | | 1 | 1 | | | | | | 1 | 1 | 1 | 1 | | | | | | 1 | | | | | | 1 | | 1 | 1 |
| Fault Diagnosis and Prognosis | 1 | | | 1 | | | | 1 | | | 1 | 1 | 1 | 1 | 1 | | | | | | | | | | | 1 | | 1 | 1 |
| Semantic Segmentation | 1 | | | 1 | 1 | | | | | | 1 | 1 | 1 | 1 | | | | | | 1 | | | | | | 1 | | 1 | 1 |
| Break Detection | 1 | | | 1 | | | 1 | | | | 1 | 1 | 1 | 1 | | | 1 | | | | | | | | | 1 | 1 | | 1 |
| Target Recognition | 1 | | | 1 | 1 | | | | | | 1 | 1 | 1 | 1 | | | | | | | 1 | | | | | 1 | | 1 | 1 |
| Night Setback Detection | | 1 | | 1 | | | | 1 | | | 1 | 1 | 1 | 1 | | 1 | | | | | | | | | | 1 | 1 | | 1 |
| Activity Identification | 1 | | | 1 | 1 | | | | | | 1 | 1 | 1 | 1 | | | | | | 1 | | | | | | 1 | | 1 | 1 |
| Fault Detection and Isolation (FDI) | 1 | | | 1 | | | | 1 | | | 1 | 1 | 1 | 1 | | | | | | | 1 | | | | | 1 | | 1 | 1 |
| Event Detection | 1 | | | 1 | | | 1 | | | | 1 | 1 | 1 | 1 | 1 | | | | | | | | | | | 1 | 1 | | 1 |
| Activity Recognition | 1 | | | 1 | 1 | | | | | | 1 | 1 | 1 | 1 | | | | | | | | | | | | 1 | | 1 | 1 |
| Signal Classification | 1 | | | 1 | 1 | | | | | | 1 | 1 | 1 | 1 | 1 | | | | | | | | | | | 1 | 1 | | 1 |
| Kick Detection | 1 | | | 1 | | | | 1 | | | 1 | 1 | 1 | 1 | | | 1 | | | | | | | | | 1 | 1 | | 1 |
| Fault Detection and Identification | 1 | | | 1 | | | | 1 | | | 1 | 1 | 1 | 1 | | | | 1 | | | | | | | | 1 | | 1 | 1 |
| Process Pattern Recognition | 1 | | | 1 | 1 | | | | | | 1 | 1 | 1 | 1 | | | | | | | | | | | | 1 | 1 | | 1 |
| Total | 80 | 19 | 18 | 117 | 40 | 30 | 19 | 10 | 9 | 9 | 117 | 64 | 53 | 117 | 64 | 9 | 7 | 6 | 6 | 6 | 5 | 5 | 4 | 3 | 2 | 117 | 66 | 51 | 117 |

**Figure 35:** Jargon-Problem Domain Matrix

Table 18: Data Source-Solution Domain Matrix

Table 19: Problem Domain-Solution Domain Matrix

Table 20: Approach-Solution Domain Matrix

Table 21: Jargon-Solution Domain Matrix